\documentclass[twoside]{article}

\usepackage[accepted]{aistats2025}
\usepackage[round]{natbib}

\usepackage{enumitem}
\setlist[itemize]{noitemsep} 
\usepackage{subcaption}
\usepackage{graphicx}
\usepackage{booktabs}
\usepackage{multirow}
\usepackage{amsthm}
\usepackage{amsmath}
\usepackage{amssymb}
\usepackage{amsfonts}
\usepackage{url}
\usepackage{datetime2}
\usepackage{hyperref}
\usepackage[dvipsnames]{xcolor}         
\usepackage{mathrsfs}
\hypersetup{
    colorlinks=true,
    linkcolor=blue,
    filecolor=magenta,      
    urlcolor=NavyBlue,
    citecolor=NavyBlue,
    filecolor=blue,
}

\newtheorem{proposition}{Proposition}
\newtheorem{theorem}{Theorem}
\newtheorem{definition}{Definition}

\newcommand{\cX}{{\mathcal{X}}}
\newcommand{\cY}{{\mathcal{Y}}}

\DeclareMathOperator*{\argmax}{arg\,max}

\newcommand{\edit}[1]{{#1}}

\newcommand{\sujay}[1]{}

\begin{document}

%

%





\runningauthor{Park, Bhatt, Zeng, Wong, Koppel, Ganesh, Walters}

\twocolumn[

\aistatstitle{Approximate Equivariance in Reinforcement Learning}

\aistatsauthor{Jung Yeon Park$^*$ \And Sujay Bhatt$^\dagger$ \And  Sihan Zeng$^\dagger$ \And Lawson L.S. Wong$^*$}

\aistatsauthor{Alec Koppel$^\dagger$ \And Sumitra Ganesh$^\dagger$ \And Robin Walters$^*$}

\aistatsaddress{$^*$Northeastern University \And  $^\dagger$J.P.Morgan AI Research}]

\begin{abstract}
  Equivariant neural networks have shown great success in reinforcement learning,
  improving sample efficiency and generalization when there is symmetry in the task. However, in many problems, only approximate symmetry is present, which makes imposing exact symmetry inappropriate. Recently, approximately equivariant networks have been proposed for supervised classification and modeling physical systems. In this work, we develop approximately equivariant algorithms in reinforcement learning (RL). We define approximately equivariant MDPs and theoretically characterize the effect of approximate equivariance on the optimal $Q$ function. We propose novel RL architectures using relaxed group and steerable convolutions and experiment on several continuous control domains and stock trading with real financial data. Our results demonstrate that the approximately equivariant network performs on par with exactly equivariant networks when exact symmetries are present, and outperforms them when the domains exhibit approximate symmetry. As an added byproduct of these techniques, we observe increased robustness to noise at test time. Our code is available at \url{https://github.com/jypark0/approx_equiv_rl}.

\end{abstract}

\section{INTRODUCTION}

Symmetry is a powerful inductive bias that can be used to improve generalization and data efficiency in deep learning.
One way to leverage symmetry is through equivariant neural networks, which are model classes constrained to respect the symmetry of a known ground truth. Equivariant neural networks have successfully been applied to image classification \citep{cohen2016group, worrall2017harmonic}, particle physics \citep{bogatskiy2020lorentz}, molecular biology \citep{satorras2021n, thomas2018tensor}, and robotic manipulation \citep{wang2022so2equivariant}. Empirical studies have demonstrated that equivariant networks require much fewer data than their standard network counterparts \citep{winkels20183d, wang2022so2equivariant}, \edit{can have fewer parameters \citep{weiler2019general, he2022neural}, }and can generalize better to unseen data \citep{wang2020incorporating, fuchs2020se}.

\begin{figure}[t]
    \centering
    \includegraphics[trim={0 35 0 60}, clip, width=0.9\columnwidth]{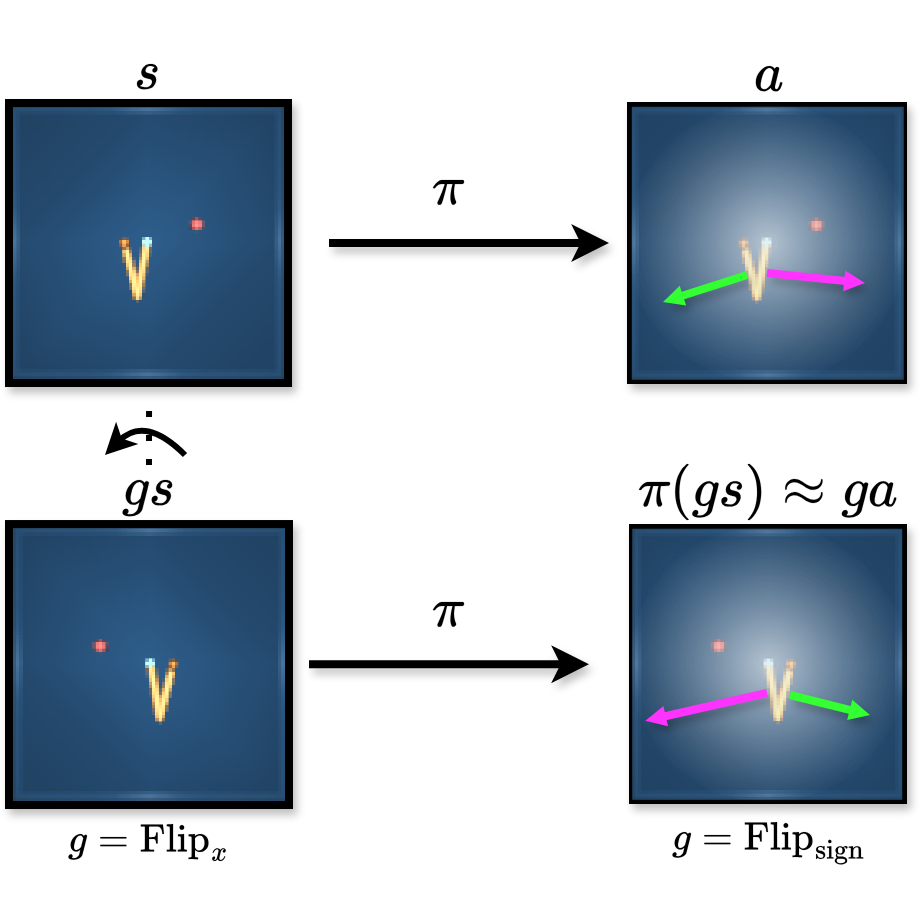}
    \caption{An approximately equivariant policy $\pi$ on a Reacher domain, where the goal is to determine the torques (green, magenta) to apply on each joint for the fingertip to reach the target (red). Due to wear, the first joint is more responsive to positive torques. When the state is flipped, the policy also flips the actions but can learn to adjust for symmetry breaking factors.}
    \label{fig:approx_equivariance}
\end{figure}

However, equivariant neural networks crucially assume that the data is perfectly symmetric in both the inputs and outputs, which may not be true in real-world data such as fluid dynamics \citep{wang2022approximately} or financial data \citep{black1986noise}. By relaxing the strict equivariance constraints, approximately equivariant networks can outperform exactly equivariant and unconstrained networks in the presence of asymmetry. While various approaches to achieve approximate equivariance have been proposed \citep{wang2022approximately, van2022relaxing, mcneela2023almost, kim2023regularizing}, they focus on vision-based tasks or dynamics modeling.

One area where symmetry has been especially useful is in reinforcement learning (RL), where equivariant networks greatly improve sample efficiency \citep{wang2022so2equivariant, zhu2022sample}, a key challenge in RL. However, most works consider exact symmetry and use exactly equivariant networks, which cannot address 
symmetry breaking in the reward or transition functions or noise in the observations.  In this work, we employ relaxed group and steerable convolutional neural networks for RL \citep{wang2022approximately}; they are flexible enough to adapt to approximate equivariance but also have improved efficiency and robustness.

In this paper, we theoretically and empirically investigate approximately equivariant reinforcement learning. 
Our key contributions are to:
\begin{itemize}
    \item 
    formalize the notion of approximately equivariant MDPs and prove the (optimal) value function in such MDPs exhibits approximate equivariance, motivating the use of approximately equivariant RL,
    \item introduce a novel approximately equivariant RL architecture using relaxed group convolutions,
    \item demonstrate improved sample efficiency and robustness to noise for our approximately equivariant RL compared to other baselines with or without symmetry biases,
    \item successfully apply approximate equivariant RL to real-world financial data.
\end{itemize}

\section{RELATED WORK}

\paragraph{Equivariant Reinforcement Learning}
Early works explored equivalence classes in reinforcement learning from the lens of abstractions by defining MDP homomorphisms \citep{ravindran2002model, zinkevich2001symmetry}. More recently, several approaches have combined function approximation with RL with equivariant neural networks \citep{van2020mdp, wang2022so2equivariant, mondal2020group} with significantly improved sample efficiency. However, all of these works considered perfectly symmetric domains where the policy is constrained to be exactly equivariant. This paper considers domains with symmetry breaking factors where exactly equivariant networks can be suboptimal.

\paragraph{Approximate Equivariant Architectures}
There has been recent interest in exploring approximate equivariance and approximately equivariant neural networks \citep{finzi2021residual, wang2022approximately, romero2022learning, van2022relaxing, mcneela2023almost, petrache2024approximation, samudre2024symmetry}. \cite{wang2022approximately, wang2024discovering} use a linear combination of exactly equivariant convolution kernels with learnable weights to achieve relaxed equivariance and discover symmetry breaking factors. \cite{van2022relaxing} define a nonstationary kernel and a tunable frequency parameter to control the amount of approximate equivariance. \cite{mcneela2023almost} propose using a neural network to approximate the exponential map from the Lie algebra to the group to learn almost equivariant functions. \cite{petrache2024approximation} give theoretical bounds on when approximate equivariance can improve generalization.
However, none of these works studied approximate equivariance in RL, the main focus of this work. 

Closest to our setting is Residual Pathway Priors \citep{finzi2021residual}, which considered soft equivariance constraints in model-free RL. They construct a relaxed equivariant neural network layer as the sum of an exactly equivariant and a non-equivariant layer with a prior on the equivariant layer. We take a different approach in this work and use relaxed group convolutions \cite{wang2022approximately}, which are flexible enough to learn different outputs for each transformation.

\paragraph{Learning with Latent Symmetry}
Other works also apply equivariant neural networks to domains with latent symmetry. These are cases where the full state has exact symmetry but only partial observations with an unknown group action are available to the model.  \cite{park2022learning} learn the out-of-plane rotations from 2D images using a symmetric embedding network while others have learned 3D rotational features from images using manifold latent variables \citep{falorsi2018explorations} or disentanglement \citep{quessard2020learning}. \cite{wang2022surprising} find that equivariant models where the group acts directly on observation space perform well in RL even with camera skew or occlusions. They define extrinsic equivariance (transformed samples are outside the data distribution) and show that it can benefit in some scenarios but can also be harmful in certain cases \citep{wang2024general}. Unlike these works where the observation is partial and does not contain full information about the state, we assume that the domains are fully observable and consider various symmetry breaking factors.


\section{BACKGROUND}
In this section, we provide some background on symmetry groups and equivariant functions. As building blocks of exactly and approximately equivariant networks, we also describe exact and relaxed group convolutions, respectively. 

\subsection{Groups and Equivariance}
\label{background_groups}
A symmetry group $G$ is a set equipped with a binary operation that satisfies associativity, existence of an identity, and existence of inverses. A group can act on vector space $X$ via a group representation $\rho_{X}$ which homomorphically assigns each element $g \in G$ an invertible matrix $\rho_X(g) \in \mathrm{GL}(X)$. 
For example, for a finite group $G$, the regular representation acts on $\mathbb{R}^{|G|}$ by permuting basis elements $\{ e_g : g \in G\}$ as $\rho_{\mathrm{reg}}(h)e_g = e_{hg}$. 
A function $f: X \rightarrow Y$, $x \mapsto y$ is $G$-equivariant if $f(\rho_{X}(g)(x)) = \rho_{Y}(g)f(x)$. That is, transformations of the input $x$ by $g$ correspond to transformations of the output by the same group element. We can enforce this constraint in equivariant neural networks to learn only over the space of equivariant functions by replacing linear layers with group or steerable convolutional layers. One benefit of enforcing equivariance is lower sample complexity as the network searches over a reduced function class. 

\subsection{Group Convolution}
One method of constructing equivariant network layers is by group convolution \citep{cohen2016group}, which we briefly describe here. Group convolutions map between features which are signals over the group $f: G \rightarrow \mathbb{R}$.  For inputs not natively of this form, a lift operation must first be performed. 
Let $\psi_\theta: G \rightarrow \mathbb{R}$ be the convolutional kernel parameterized by $\theta$. A $G$-equivariant group convolutional layer is defined as
\begin{align}
    (f \star \psi_\theta)(g) = \sum_{h \in G} \psi_\theta(g^{-1} h)f(h).
\end{align}
Equivariance follows from the fact that the kernel depends only on the product $g^{-1}h$ and not the specific elements $(g,h)$. For example, if we consider equivariance across translations, we obtain the standard convolution where $h,g \in \mathbb{Z}^2$ and $g^{-1}h= h - g$. Another possible approach to constructing equivariant network layers is with $G$-steerable convolutions \citep{cohen2017steerable}, which can generalize to continuous groups.


\subsection{Relaxed Group Convolution}
A key component of our method is the relaxed version of the group convolution \citep{wang2022approximately}. The kernel $\psi$ is replaced with several kernels $\{\psi_l\}_{l=1}^{L}$ and the output is composed as a linear combination. The relaxed group convolution is defined as
\begin{align}
    (f \widetilde{\star} \psi_\theta)(g) = \sum_{h \in G} f(h) \sum_{l=1}^{L} w^l(h) \psi_\theta^{l}(g^{-1}h),
\end{align}
where $w^l$ are the relaxed weights and each $\psi_\theta^{l}$ are constrained to be exactly equivariant.
Note that as $w^l(h)$ depends on the specific element $h$, this breaks the strict equivariance of the group convolution.  
\cite{wang2022approximately} also introduce relaxed versions of steerable convolutions, see \cite{wang2022approximately} or Appendix~\ref{app:steerable_conv} for more details.

\subsection{Approximate Equivariance}
There have been several different definitions of approximate, relaxed, or partial equivariance. In this paper, we use the definition given by \cite{petrache2024approximation}. We give some background to build up to the definition. Let $G$ be a group and $f: X \rightarrow Y, x \mapsto y$ be the task function.
\begin{definition}[Equivariance Error]
For $g \in G$ and $x \in X$, the equivariance error $ee(f, g, x)$ is defined as 
\begin{align}
ee(f, g, x) = \| f(g(x)) - g(f(x)) \|,
\end{align}
\end{definition}
Equivariance error measures exactly how far a function is from perfect equivariance with respect to $G$ for a particular $x$. 
For an exactly $G$-equivariant function, $ee(f,g,x) = 0$ for all $g \in G$ and  $x \in X$. 

\begin{definition}[$\varepsilon$-stabilizer]
The $\varepsilon$-stabilizer of $f$ and $G$ is defined as
\begin{align}
\textrm{Stab}_{\varepsilon}(f, G) = \{g \in G \mid ee(f, g, x) \leq \varepsilon \}.
\end{align}
\end{definition}
The $\varepsilon$-stabilizer gives the set of group elements for which the equivariance error is under some threshold.

\begin{definition}[Approximate $G$-Equivariance] 
Given a function $f: \cX \to \cY$ and a group $G$, $f$ is approximately $G$-equivariant if $\textrm{Stab}_{\varepsilon}(f,G) = G$.
\end{definition}
We adopt the definition of approximate equivariance where $f$ has bounded equivariance error for all $g \in G$, in contrast to \textit{partial equivariance}, where $\textrm{Stab}_{\varepsilon}(f,G) < G$.

\section{METHOD: APPROXIMATELY EQUIVARIANT REINFORCEMENT LEARNING}

We first theoretically characterize the problem by defining approximately equivariant Markov decision processes (MDP).  We then prove that environments with approximate symmetry admit approximately invariant $Q$ functions. This motivates our method of using approximately equivariant neural networks to learn the policy and $Q$ function.    



\subsection{Approximately Equivariant MDP}
Consider an infinite-horizon discounted-reward Markov decision process (MDP) represented by a tuple $M=(\mathcal{S}, \mathcal{A}, P, R, \gamma)$ with state space~$\mathcal{S}$, action space~$\mathcal{A}$, instantaneous reward function~$R : \mathcal{S} \times \mathcal{A} \rightarrow \mathbb{R}$, a transition function~$P: \mathcal{S} \times \mathcal{A}  \rightarrow \Delta_{\mathcal{S}}$ and discount factor~$\gamma \in (0,1)$. 

Let $\pi : \mathcal{S} \rightarrow \Delta_\mathcal{A}$ be a policy giving the probability $\pi(a|s)$ of taking action $a$ in state $s$. The expected cumulative reward of using the policy starting from state $s$ (or state $s$ and action $a$) are the value functions defined as follows
\begin{align}
\begin{gathered}
V^{\pi}(s) := \mathbb{E}^{\pi}\Big[ \sum_{k=0}^{\infty} \gamma^{k} R(s_k,a_k) \Big| s_0 = s \Big],\\
Q^{\pi}(s,a) := \mathbb{E}^{\pi}\Big[ \sum_{k=0}^{\infty} \gamma^{k} R(s_k,a_k) \Big| s_0 = s, a_0=a \Big].
\end{gathered}
\label{eq:def_V_Q}
\end{align}

The goal is to find a policy $\pi^{*}$ that maximizes the expected return with an initial state distribution $\xi$
\[\pi^{*}:=\argmax_\pi \mathbb{E}_{s_0\sim\xi}[V^{\pi}(s_0)].\]
We denote $V^{*}=V^{\pi^{*}}$ and $Q^{*}=Q^{\pi^{*}}$.

Let $G$ be a group acting on $\mathcal{S}$ and $\mathcal{A}$. Denote the action of an element $g \in G$ on $s$ and $a$ by $gs$ and $ga$, respectively. 
We now extend the definition of Equivariant MDPs \citep{van2020mdp} to cases where the symmetry is approximate.
\begin{definition}  \label{def:G_inv}
An MDP is $(G, \epsilon_R, \epsilon_P)$-invariant if
\begin{align*}
    &| R(gs, ga) - R(s, a) | \leq \epsilon_R, \forall g \in G \\
    & d_{\mathscr{F}} \Big( P(gs' \mid gs, ga) , P(s' \mid s, a) \Big)  \leq \epsilon_P, \forall g \in G,
\end{align*}
where~$d_{\mathscr{F}}(\mu,\nu):= \sup_{f \in \mathscr{F}} \Big| \int_{\mathcal{S}} f d\mu - \int_{\mathcal{S}} f d\nu \Big|$ is an integral probability metric (IPM) between two distributions~$\mu, \nu \in \Delta(\mathcal{X})$.
\end{definition}
Some well known examples of IPM include~\citep{sriperumbudur2009integral}: total variation distance ($\mathscr{F} = \{ f: \| f\|_{\infty} \leq 1 \}$) and  Kantorovich metric ($\mathscr{F} = \{ f: \| f\|_{\text{Lip}} \leq 1 \}$). A useful property of IPMs is, given a function class~$\mathscr{F}$ and a function~$f$~\citep{muller1997does}
\[
\Big| \int_{\mathcal{S}} f d\mu - \int_{\mathcal{S}} f d\nu \Big| \leq \rho_{\mathscr{F}}(f) \cdot d_{\mathscr{F}}(\mu,\nu),
\]
where the Minkowski functional  w.r.t~$\mathscr{F}$ is 
\[
\rho_{\mathscr{F}}(f) = \inf \{\rho \in \mathbb{R}_{\geq 0} : \rho^{-1} f \in \mathscr{F}\}.
\]
For the total variation distance~$\rho_{\mathscr{F}}(f):= \frac{1}{2}(\max{f} - \min{f})$
and for Kantorovich metric~$\rho_{\mathscr{F}}(f):= \|f\|_{\text{Lip}}$.

The following theorem provides a characterization of the gap between the value functions in the original and symmetry transformed domain, for the $(G, \epsilon_R, \epsilon_P)-$invariant MDP described in~Definition~\ref{def:G_inv}. Theorem~\ref{thm:Gap_Q} highlights that the Q-function is approximately group-invariant, where the approximation is now a function of the reward and transition mismatch, the discount factor, and the Minkowski functional evaluated on the optimal value function.

\begin{theorem} \label{thm:Gap_Q}
Let the rewards $R$ be bounded $R_{\min} \leq R \leq R_{\max}$, $0 \leq \gamma <1$ and let~$g \in G$ be an onto mapping. 
For any state~$s$ and action~$a$, we have
\begin{gather*}
|Q^*(s,a) - Q^*(gs,ga)| \leq \alpha,\\
|V^*(s) - V^*(gs)| \leq \alpha,
\end{gather*}
where~$\alpha = \frac{\epsilon_R + \gamma \rho_{\mathscr{F}}(V^*) \epsilon_P}{1 - \gamma}$.
\end{theorem}
Theorem~\ref{thm:Gap_Q} implies that when the invariance mismatch is small -- i.e.,~when the domain has only minor symmetry violations -- the Q-function is approximately group-invariant. A proof is provided in Appendix~\ref{app:proofs}. 
Note that, in Theorem~\ref{thm:Gap_Q}, when the Kantorovich metric is used for uncertainty characterization, $\rho_{\mathscr{F}}(V^*) = \|V^* \|_{\text{Lip}}$, where~$\|\cdot\|_{\text{Lip}}$ is the Lipschitz norm of the value function~\citep{gelada2019deepmdp}. For total variation distance, $\rho_{\mathscr{F}}(V^*) =|R_{\max} - R_{\min}|$.

Also, from Theorem~\ref{thm:Gap_Q}, it is clear that when~$\gamma \in [0,1)$, we obtain a non-trivial characterization, while~$\gamma=1$ results in a trivial and uninformative bound. This is the limitation of the infinite horizon setting, and can be remedied by considering an arbitrary finite-horizon setup. We do this for the sake of completeness in Appendix~\ref{sec:Ap_nodisc}. We not only show that the finite horizon setup allows for a time-dependent transition function, but also obtain an approximate group-invariance of the time dependent Q-function in terms of similar elements that appear in Theorem~\ref{thm:Gap_Q}.  

\edit{
There are different ways to use the above results. One can discover how approximate the value functions are and learn $\alpha$, or one can incorporate approximate equivariance into the model and leverage the benefits of equivariance. We take the latter approach and consider approximately equivariant networks for the policy and critic in domains with inexact symmetry.
}





\begin{figure*}[t]
    \centering
    \includegraphics[trim={0 100 0 0},clip, width=0.9\textwidth]{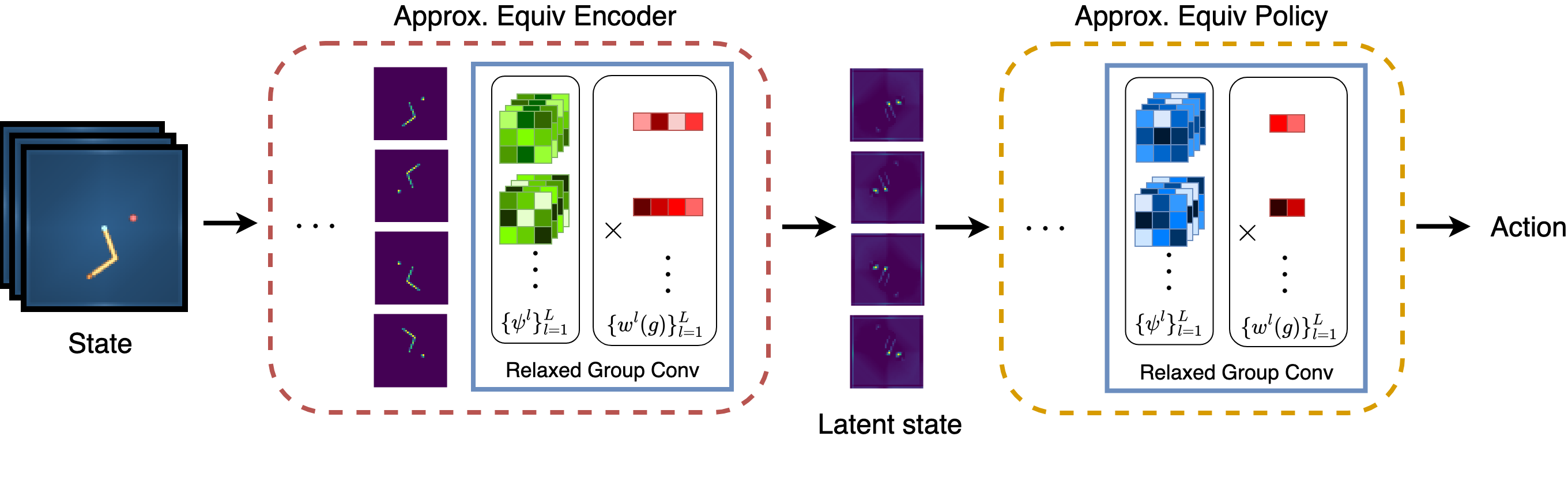}
    \caption{Illustration of the approximately $D_2$-equivariant encoder and policy (critic is not shown for space). The $D_2$ group consists of vertical reflections and $\pi$ rotations. Both the encoder and policy consist of relaxed group convolution layers.}
    \label{fig:architecture}
    \vspace{-0.3cm}

\end{figure*}

\begin{figure*}[t]
    \centering
    \begin{subfigure}[b]{\textwidth}
         \centering
         \includegraphics[width=\textwidth]{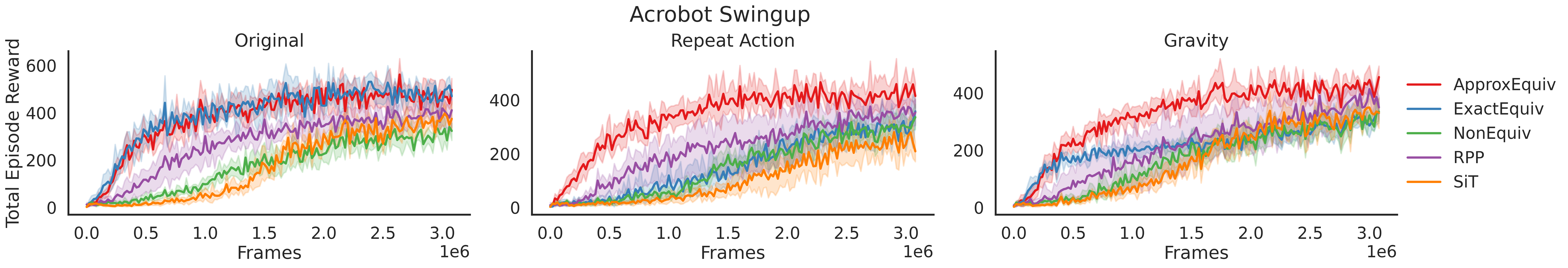}
     \end{subfigure}
    \begin{subfigure}[b]{\textwidth}
         \centering
         \includegraphics[width=\textwidth]{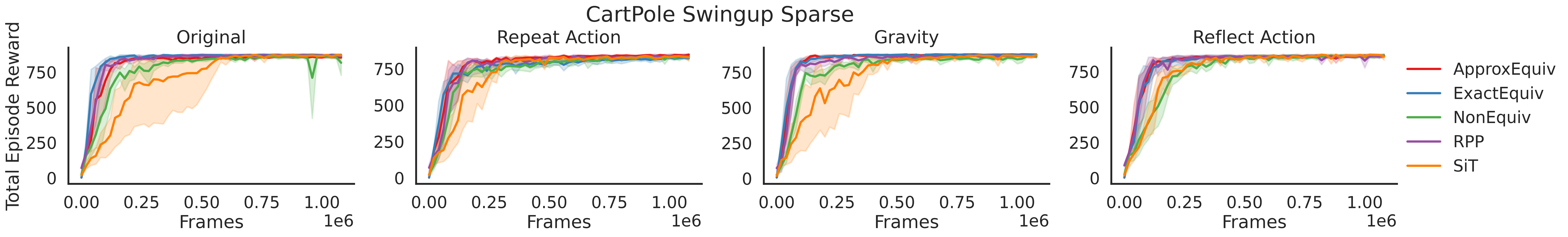}
     \end{subfigure}
    \begin{subfigure}[b]{\textwidth}
         \centering
         \includegraphics[width=\textwidth]{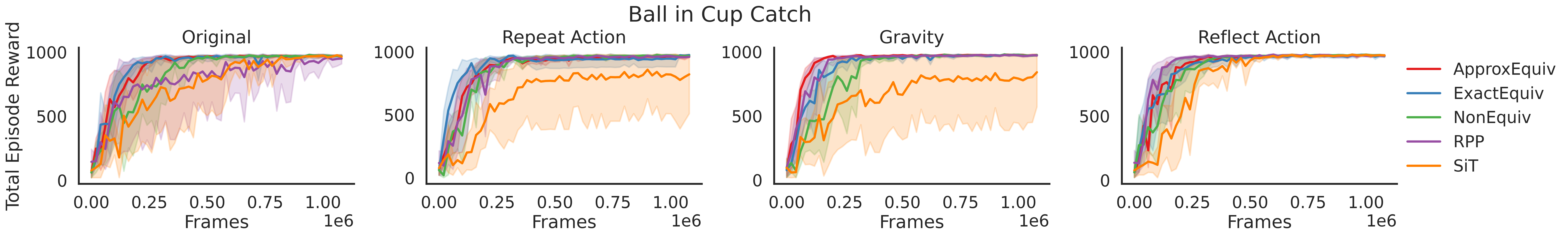}
    \end{subfigure}
    \begin{subfigure}[b]{\textwidth}
         \centering
        \includegraphics[width=\textwidth]{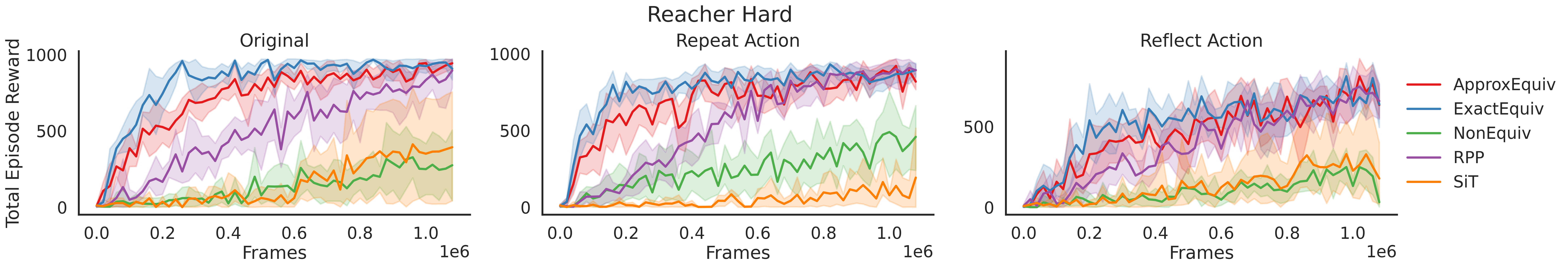}
    \end{subfigure}
    \caption{Total episode reward on selected domains in the DeepMind Control Suite, shaded regions indicate $95\%$ confidence intervals (CI). \edit{Compared to an exactly equivariant agent (ExactEquiv), our approximately equivariant agent (ApproxEquiv) outperforms in \texttt{Acrobot}, performs similarly in two domains, and is slightly worse in the \texttt{Reacher} domain. ApproxEquiv can outperform ExactEquiv on some modified variants with inexact symmetry as it can adjust for symmetry breaking. Our agent outperforms all other baselines, including a non-equivariant agent, suggesting that relaxed symmetry is a good inductive bias.}
    }
    \label{fig:dmc_curves}

\end{figure*}

\subsection{Approximately Equivariant Actor-Critic}
We propose approximately equivariant versions of two commonly used actor-critic algorithms, DrQv2 \citep{yarats2021mastering} and SAC \citep{haarnoja2018soft}.
In doing so, we generalize exactly equivariant versions of SAC \citep{wang2022so2equivariant} and DrQv2 \citep{wang2022surprising} from previous works by replacing strictly equivariant layers with relaxed equivariant layers. 

\paragraph{Illustrative Example}
We \edit{first }illustrate how to apply our proposed approximately equivariant actor-critic architecture on the \texttt{Reacher} domain; see Figure~\ref{fig:architecture}. \edit{The objective is to actuate a two-joint arm so that the end effector reaches the red point.} 
The state is a stack of consecutive images $s \in \mathbb{R}^{C \times H \times W}$ and the action $a \in \mathbb{R}^2$ corresponds to torques for the first and second arms. \edit{There is clear rotational and reflectional symmetry in this domain. If the state (image) is rotated, the action should be invariant to rotations as they are angular torques. If the state is reflected, then the action would also correspondingly be flipped (in sign). However, as in the example in Figure~\ref{fig:approx_equivariance}, the first joint is more responsive to positive torques, which breaks rotational and reflectional symmetry. }

For this domain, we implement approximate equivariance to the group $D_2$ of vertical reflections and $\pi$ rotations. The group $D_2$ transforms the input states by image transformations, \edit{where the input images are reflected or rotated}.  Latent representations are images $z \colon \mathbb{R}^2 \to \mathbb{R}^C$ where $g \in D_2$ acts on the pixel axes by image transformation and on the channel axis by permutations corresponding to the regular representation of $D_2$, i.e. $(gz)(x,y) = \rho_{\mathrm{reg}}(g) z(g^{-1} \cdot (x,y))$. \edit{Note that the latent representations can be high-dimensional, consisting of a direct sum of several different or repeated low-dimensional representations of $D_2$.}
%
%
For the output, the torques $a_1$ and $a_2$ are scalars that change sign under reflection but are invariant under rotations. 

\paragraph{Encoder, Policy, and Critic}
\edit{We extend exactly equivariant versions of SAC \citep{wang2022so2equivariant} and DrQv2 \citet{wang2022surprising} by replacing }each group convolution with relaxed group convolutions 
for the encoder, policy, and critics. Practically, each relaxed group convolution layer contains $L$ exactly equivariant kernels $\psi_l$ and  
the output is a linear combination of the outputs of these convolutions and relaxed weights $w^l(g)$. The $w^l(g)$ also transform as the regular representation of $G$, see Section~\ref{background_groups} for the definition for finite groups. 

The encoder $E$ and the policy $\pi$ are approximately equivariant. The latent state $z$ output by $E$ is defined to transform as the \edit{direct sum of }regular representations of $G$. The action representation is domain-specific. 
The critics are approximately invariant and output scalars $q_{(s,a)}$ that are fixed by $G$, i.e. transform via the trivial representation. For more details, please see Section~\ref{sec:experiments} and Appendix~\ref{app:method}. 

\edit{In the case of continuous groups, we can also construct relaxed steerable versions of the encoder, policy, and critics. Analogous to the group convolution case, we can replace the exactly equivariant steerable convolutions with relaxed steerable convolutions. See Appendix~\ref{app:method} for more details.}

\begin{figure}[t]
    \centering
    \begin{subfigure}[b]{0.24\columnwidth}
         \centering
         \includegraphics[width=\textwidth]{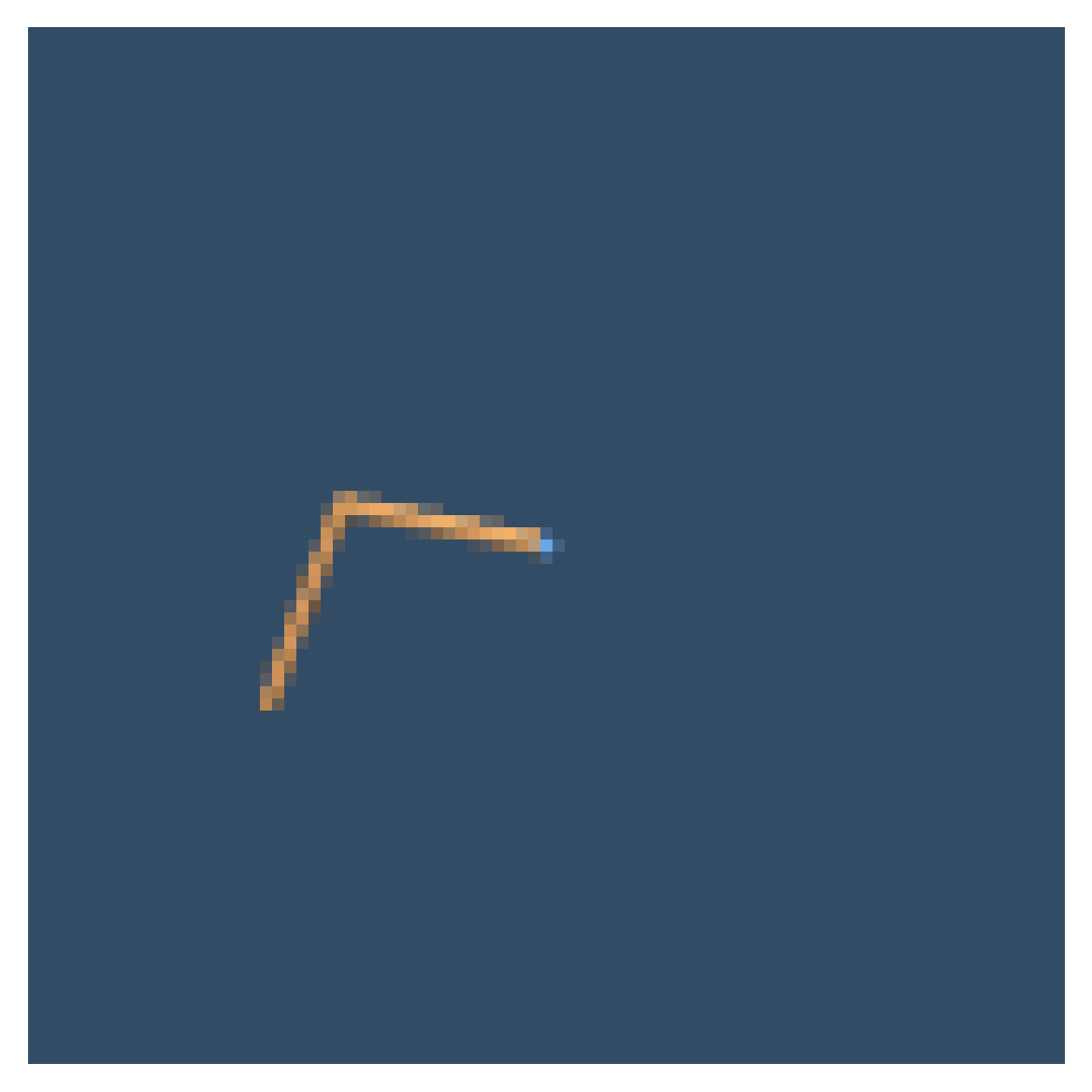}
     \end{subfigure}
    \begin{subfigure}[b]{0.24\columnwidth}
         \centering
         \includegraphics[width=\textwidth]{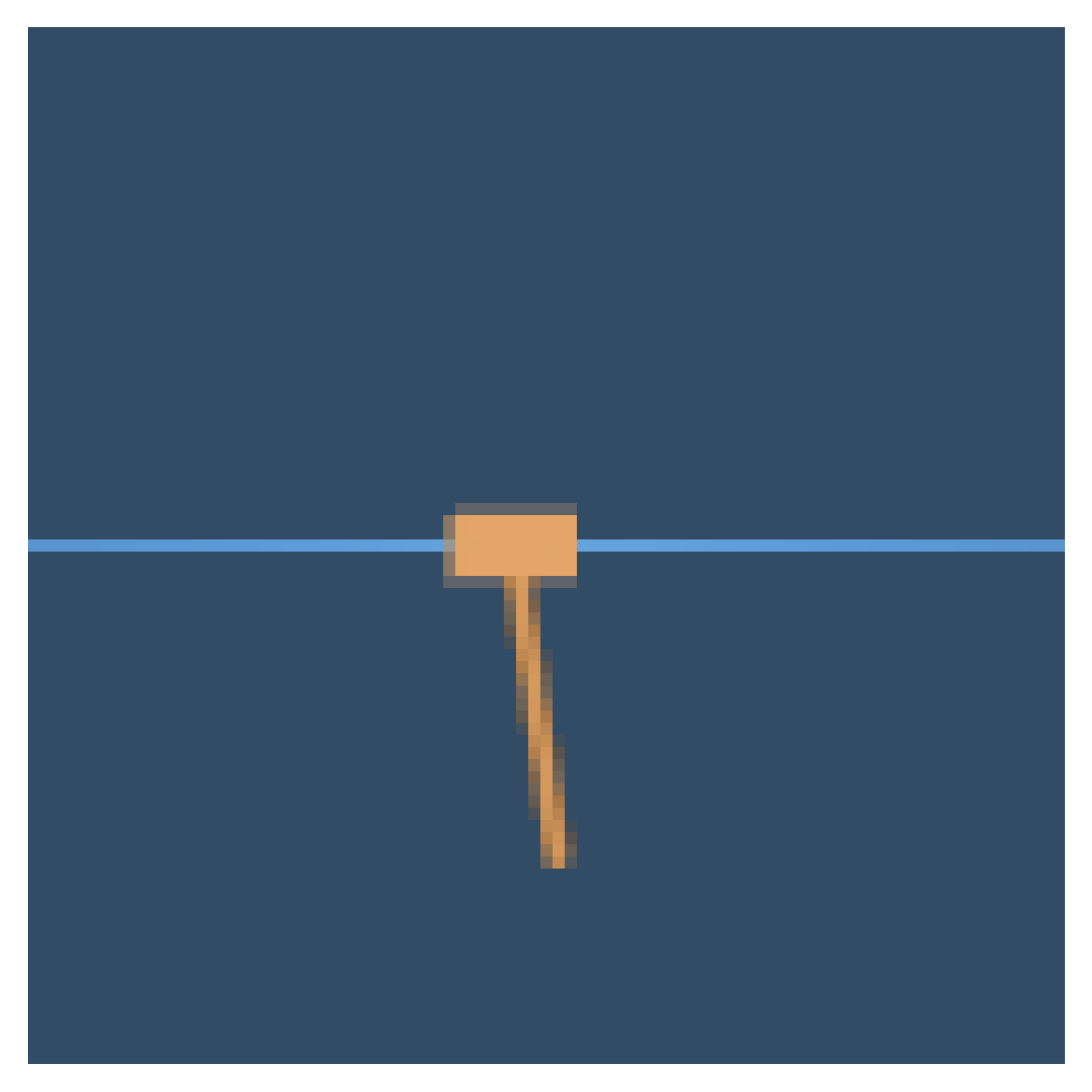}
     \end{subfigure}
    \begin{subfigure}[b]{0.24\columnwidth}
         \centering
         \includegraphics[width=\textwidth]{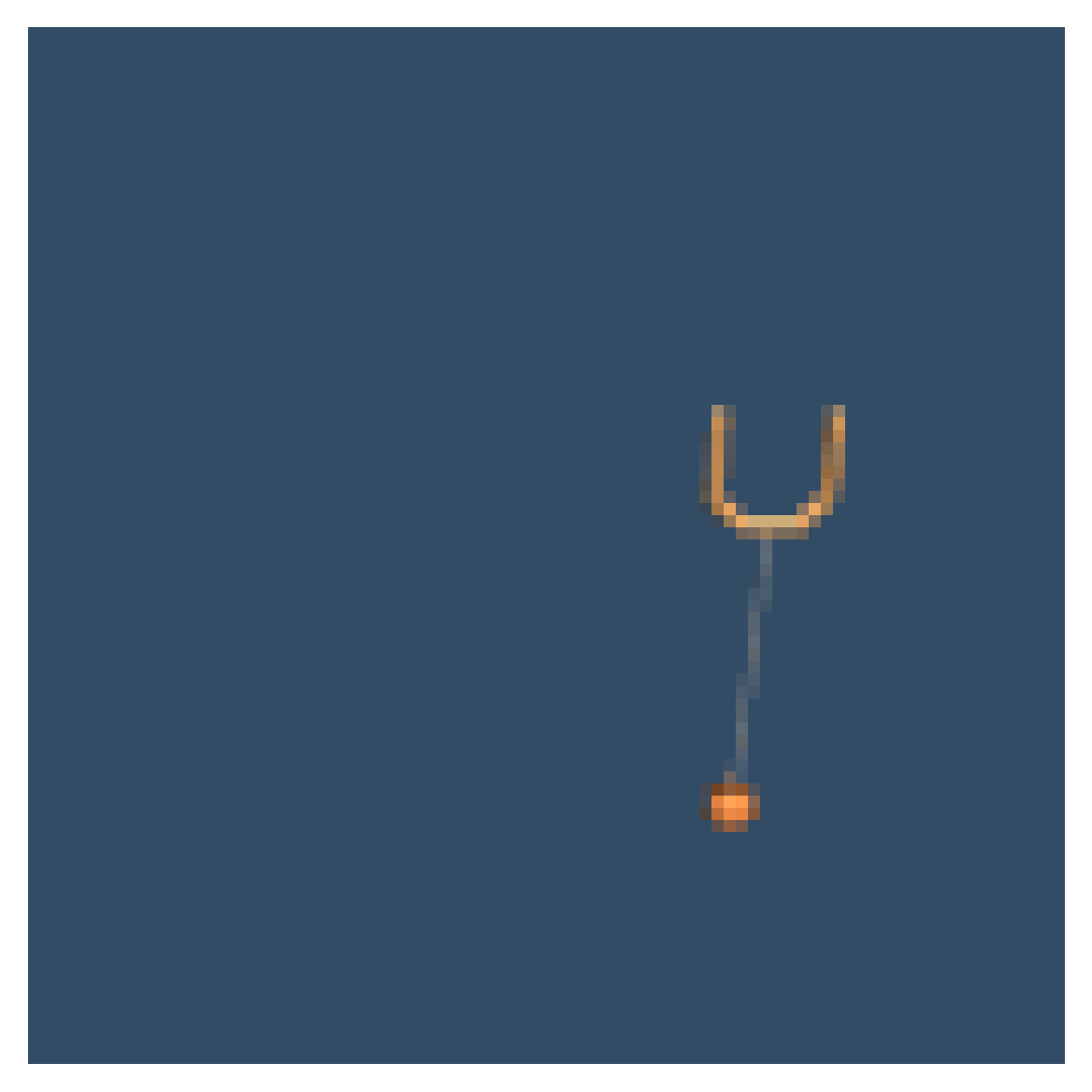}
    \end{subfigure}
    \begin{subfigure}[b]{0.24\columnwidth}
         \centering
        \includegraphics[width=\textwidth]{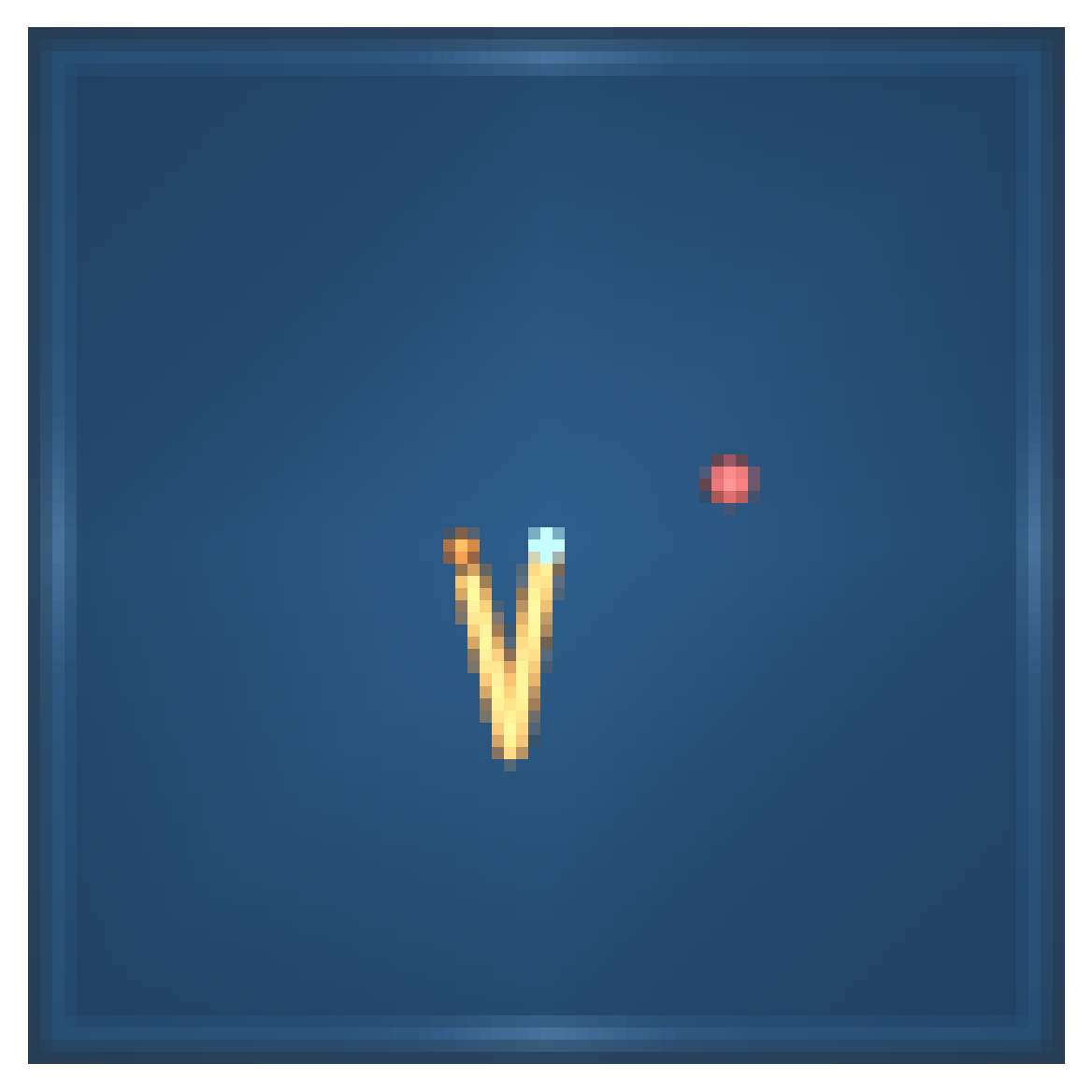}
    \end{subfigure}
    \caption{Selected domains in DeepMind Control Suite. The domains were modified to remove extrinsic symmetry and to include several types of symmetry breaking factors such as repeating or reflecting actions in certain states, or by modifying gravity.}
    \label{fig:dmc_domains}
    \vspace{-0.3cm}
\end{figure}

\section{EXPERIMENTS}
\label{sec:experiments}

{\sujay{We should include a section on symmetry breaking.. constructing realistic datasets that have approximate symmetry}}

We experiment on how approximately equivariant RL compares to methods with exact equivariance and no equivariance in domains with both exact symmetry and various symmetry breaking factors, and to elucidate when approximate equivariance should be preferred. We consider standard continuous control domains and stock trading with real-world data.

\begin{table*}[t]
\centering
\caption{Total episode reward on $50$ rollouts for the best policy in the original and noisy domains. Gray values indicate $95\%$ CI. \texttt{ApproxEquiv} learns a better policy than baselines on the modified domains and is more robust to noisy inputs.}
\resizebox{\textwidth}{!}{%
\begin{tabular}{@{}ll|rrr|rrr@{}}
\toprule
 & \multicolumn{1}{c|}{} & \multicolumn{3}{c|}{No Noise} & \multicolumn{3}{c}{Noisy} \\
 & \multicolumn{1}{c|}{} & \multicolumn{1}{c}{ApproxEquiv} & \multicolumn{1}{c}{ExactEquiv} & \multicolumn{1}{c|}{NonEquiv} & \multicolumn{1}{c}{ApproxEquiv} & \multicolumn{1}{c}{ExactEquiv} & \multicolumn{1}{c}{NonEquiv} \\ \midrule
\multirow{2}{*}{\sc{Acrobot}} & Original & $389 {\color{gray}\scriptstyle \pm 11}$ & $\textbf{522} {\color{gray}\scriptstyle \pm 21}$ & $309 {\color{gray}\scriptstyle \pm 22}$ & $344 {\color{gray}\scriptstyle \pm 14}$ & $\textbf{402} {\color{gray}\scriptstyle \pm 22}$ & $190 {\color{gray}\scriptstyle \pm 14}$ \\
 & Gravity & $\textbf{471} {\color{gray}\scriptstyle \pm 17}$ & $382 {\color{gray}\scriptstyle \pm 15}$ & $358 {\color{gray}\scriptstyle \pm 23}$ & $\textbf{369} {\color{gray}\scriptstyle \pm 15}$ & $218 {\color{gray}\scriptstyle \pm 10}$ & $202 {\color{gray}\scriptstyle \pm 12}$ \\ \midrule
\multirow{2}{*}{\sc{Cartpole}} & Original & $876 {\color{gray}\scriptstyle \pm 0.2}$ & $\textbf{881} {\color{gray}\scriptstyle \pm 0.1}$ & $\textbf{881} {\color{gray}\scriptstyle \pm 0.1}$ & $778 {\color{gray}\scriptstyle \pm 22}$ & $\textbf{855} {\color{gray}\scriptstyle \pm 0.6}$ & $572.5 {\color{gray}\scriptstyle \pm 25}$ \\
 & Repeat Action & $\textbf{859} {\color{gray}\scriptstyle \pm 0.4}$ & $749 {\color{gray}\scriptstyle \pm 13}$ & $855 {\color{gray}\scriptstyle \pm 0.6}$ & $\textbf{624} {\color{gray}\scriptstyle \pm 6.0}$ & $523 {\color{gray}\scriptstyle \pm 21}$ & $192 {\color{gray}\scriptstyle \pm 5.0}$ \\ \midrule
\multirow{2}{*}{\sc{Ball in Cup}} & Original & $961 {\color{gray}\scriptstyle \pm 0.0}$ & $958 {\color{gray}\scriptstyle \pm 0.0}$ & $\textbf{970} {\color{gray}\scriptstyle \pm 0.0}$ & $\textbf{882} {\color{gray}\scriptstyle \pm 7.7}$ & $783 {\color{gray}\scriptstyle \pm 24}$ & $0 {\color{gray}\scriptstyle \pm 0.0}$ \\
 & Gravity & $\textbf{969} {\color{gray}\scriptstyle \pm 0.0}$ & $966 {\color{gray}\scriptstyle \pm 0.0}$ & $959 {\color{gray}\scriptstyle \pm 0.0}$ & $\textbf{888} {\color{gray}\scriptstyle \pm 13.2}$ & $0 {\color{gray}\scriptstyle \pm 0.0}$ & $1.8 {\color{gray}\scriptstyle \pm 1.8}$ \\ \midrule
\multirow{2}{*}{\sc{Reacher}} & Original & $903 {\color{gray}\scriptstyle \pm 33}$ & $\textbf{950} {\color{gray}\scriptstyle \pm 15}$ & $519 {\color{gray}\scriptstyle \pm 68}$ & $\textbf{778} {\color{gray}\scriptstyle \pm 41}$ & $745 {\color{gray}\scriptstyle \pm 44}$ & $247 {\color{gray}\scriptstyle \pm 52}$ \\
 & Reflect Action & $\textbf{757} {\color{gray}\scriptstyle \pm 55}$ & $707 {\color{gray}\scriptstyle \pm 59}$ & $243 {\color{gray}\scriptstyle \pm 58}$ & $\textbf{659} {\color{gray}\scriptstyle \pm 42}$ & $217 {\color{gray}\scriptstyle \pm 41}$ & $82 {\color{gray}\scriptstyle \pm 29}$ \\ \bottomrule
\end{tabular}
}
\label{tab:dmc_test}
\end{table*}

\begin{figure*}[t]
    \centering
    \begin{subfigure}[b]{0.49\textwidth}
         \centering
         \includegraphics[width=\textwidth]{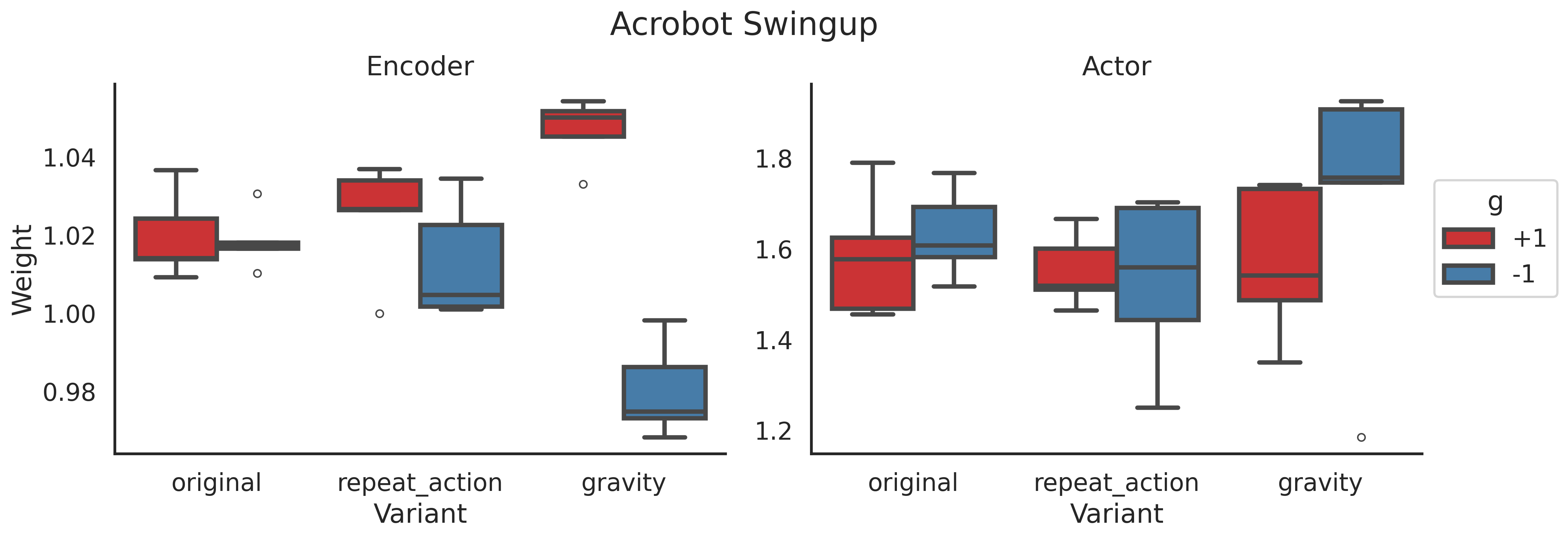}
     \end{subfigure}
    \begin{subfigure}[b]{0.49\textwidth}
         \centering
         \includegraphics[width=\textwidth]{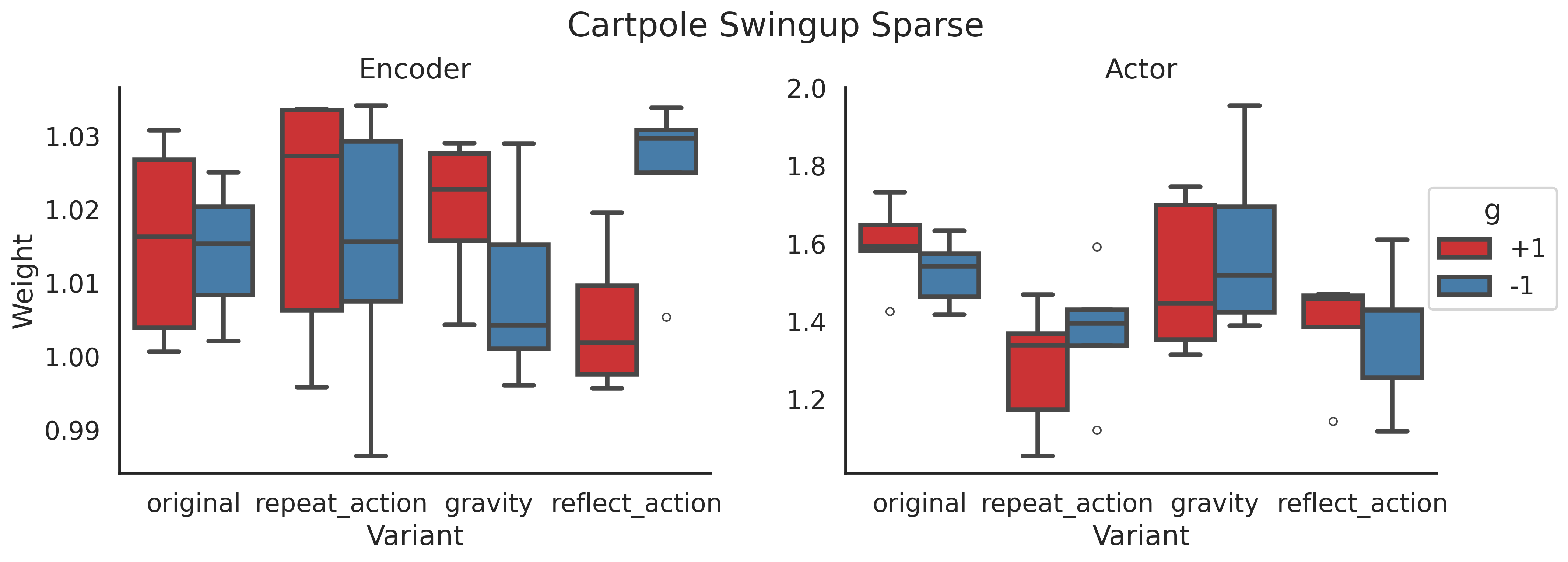}
     \end{subfigure}
    \begin{subfigure}[b]{0.49\textwidth}
         \centering
         \includegraphics[width=\textwidth]{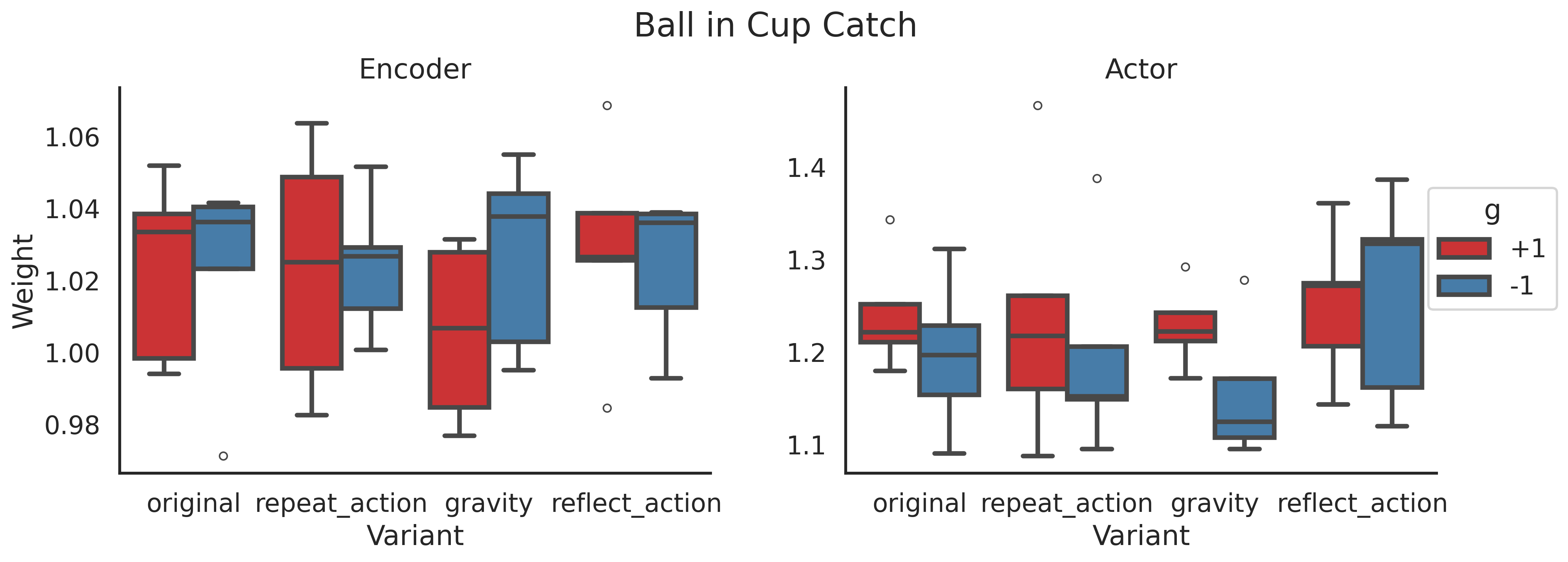}
    \end{subfigure}
    \begin{subfigure}[b]{0.49\textwidth}
         \centering
        \includegraphics[width=\textwidth]{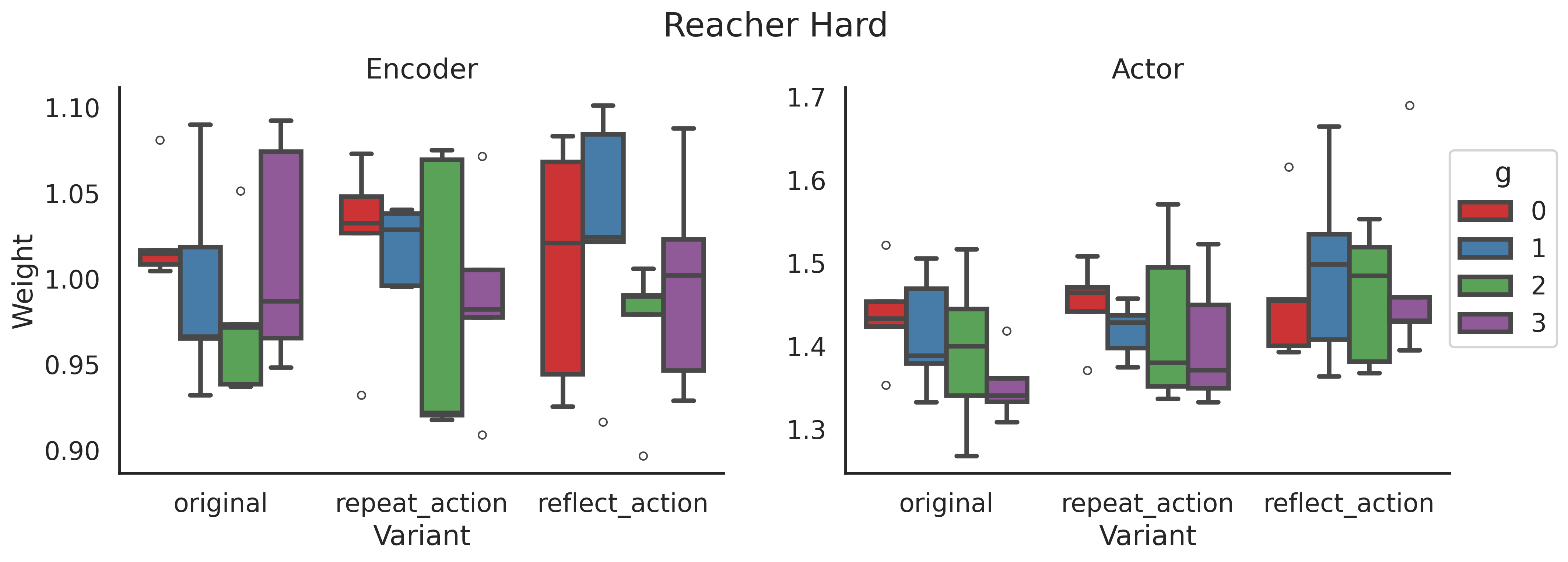}
    \end{subfigure}
    \caption{Visualization of relaxed weights for the first layer of the encoder and policy over all runs. Similar weights for each $g$ indicate perfect equivariance while differing values indicate symmetry breaking. The modified variants of most domains exhibit larger differences or increased variance in the relaxed weights compared to the \texttt{original} variant. 
    }
    \label{fig:dmc_weights}
    \vspace{-0.3cm}
\end{figure*}

\subsection{Continuous Control}
We first experiment on four continuous control domains in DeepMind Control Suite \citep{tassa2018deepmind}. 
Similar to \cite{wang2022surprising},
we consider a subset of the domains which have apparent symmetry. \texttt{Acrobot}, \texttt{Cartpole}, and \texttt{BallInCup} have reflectional symmetry described by the group $D_1$ and \texttt{Reacher} has $D_2$ symmetry.  
For all domains, the observations are a stack of $3$ consecutive RGB images.

We modify the domains to carefully control the type and degree of symmetry breaking that is present. 
We first remove fixed background features such as random stars in the sky and checkered floors (see Figure~\ref{fig:dmc_domains}).  These features break symmetry to some extent since they do not transform with the underlying state, but give a form of mild symmetry breaking termed extrinsic equivariance, which has an inconsistent impact on equivariant models \citep{wang2022surprising}.  We then introduce several different symmetry breaking factors for each domain: 1) \texttt{repeat\_action} - the action is repeated twice in a certain region of the domain, 2) \texttt{gravity} - gravity is modified from the force vector $(0, 0, -9.81)$ to $(a, -a, -9.81)$ where $a \neq 0$, and 3) \texttt{reflect\_action} - the action direction is flipped in certain regions of the domain. \texttt{repeat\_action} and \texttt{reflect\_action} test local symmetry breaking factors, while \texttt{gravity} tests a global symmetry breaking factor. See Appendix~\ref{app:dmc} for more details.

\paragraph{Models} For the continuous control tasks, we implement an approximately equivariant($\texttt{ApproxEquiv}$) version of a SOTA image-based RL algorithm DrQv2 \citep{yarats2021mastering}. We compare with exactly equivariant ($\texttt{ExactEquiv}$) and non equivariant ($\texttt{NonEquiv}$) versions of the same architecture.  We largely use the hyperparameters from \cite{yarats2021mastering} but reduce the latent dimension for more tractable computation for all methods. \edit{We also compare against an approximately equivariant model, Residual Pathway Priors (RPP) \citep{finzi2021residual}, and a self-supervised symmmetry-aware model, SiT \citep{weissenbacher2024sit}. We extend RPP to the DrQv2 architecture by using RPP layers in the encoder, policy, and critics. We find that RPP is somewhat sensitive to the speed $\tau$ of the critic moving average (as mentioned in the original paper), and had to reduce its value for $\texttt{Acrobot}$ and $\texttt{BallInCup}$ for stability. We also extend SiT to the DrQv2 architecture by using an SiT as the encoder and standard MLPs for the policy and critics. Although we adapted the code from the official SiT implementation, we were unable to modify the input image sizes and had to use the image size used in the original paper (64px).}
%
%

\paragraph{Results}

Figure~\ref{fig:dmc_curves} show the total episode reward over training. As expected, we confirm that \texttt{NonEquiv} has much lower sample efficiency than the models with a symmetry bias. In the \texttt{repeat\_action} and \texttt{reflect\_action} variants of Acrobot, \texttt{ApproxEquiv} significantly outperforms \texttt{ExactEquiv} and \texttt{RPP}. It does slightly worse than \texttt{ExactEquiv} on the \texttt{Reacher} domain but beats \texttt{RPP}, suggesting that the symmetry breaking we introduced was not strong enough to achieve incorrect equivariance. It is also possible that \texttt{ExactEquiv} 
can infer the symmetry breaking factors from the 3 frames of input, making the task a case of extrinsic equivariance where an equivariant model can succeed \cite{wang2022surprising}. In \texttt{CartPole} and \texttt{BallInCup}, all methods perform similarly and learn an optimal policy quickly. In domains with exact symmetry (\texttt{original}), our method \texttt{ApproxEquiv} performs similarly to \texttt{ExactEquiv}, showing there is no cost in performance by giving the model the ability to adapt to symmetry breaking in cases where it is not needed. This result supports Proposition 3.1 from \cite{wang2024discovering}, which proves that relaxed group convolutions initialized to be exactly equivariant stay exactly equivariant when trained with exact data symmetry. 


We visualize the relaxed weights of the first layers of the encoder and policy over all runs in Figure~\ref{fig:dmc_weights}.  If these weights are equal, the model is equivariant; the more they differ the more the model has relaxed the symmetry constraint.  For \texttt{Acrobot} and \texttt{CartPole}, the weights differ more for the modified domains than the original symmetric domain, especially for the encoder, while the policy weights vary more for the modified domains of \texttt{BallInCup}. This indicates the relaxed equivariant models have adapted to the symmetry breaking in the domains.

To quantitatively evaluate the models, we select the best-performing policy from all runs and measure the total reward over $50$ episodes. The results echo the training curves in Figure~\ref{fig:dmc_curves}, where \texttt{ApproxEquiv} performs well, particularly in the domains with symmetry breaking factors (see Table \ref{tab:dmc_test}). 

To test whether approximately equivariant models are robust to noisy observations, we also consider variants of the domains where Gaussian noise are added to the input images only at test time ($\sigma=0.02$ for \texttt{Acrobot} and \texttt{Reacher}, $\sigma=0.06$ for \texttt{CartPole} and \texttt{BallInCup}). Interestingly, we find that our approach is more robust to noisy inputs than \texttt{ExactEquiv} or \texttt{NonEquiv}, especially on the \texttt{BallInCup} and \texttt{Reacher} domains. \edit{We further experiment with \textit{training} on noisy data and test on noisy domains to see which policies are more robust, see Table~\ref{tab:noisy_training} in Appendix~\ref{app:experiments}. We find that in the \texttt{BallInCup} domains, the approximately equivariant agent is still more robust to noise than the fully equivariant or non equivariant baselines.}

\subsection{Stock Trading} We also consider a stock trading task using real world price data, formulated as an MDP \citep{liu2018practical}. Given a fixed amount of initial cash, the objective is to learn the optimal number of stocks to buy and sell (once daily) to maximize the portfolio value. The state consists of the current cash balance, the stock prices, the number of shares in the current portfolio, and other technical indicators of each stock. The actions are the number of stocks to buy and sell for each stock. The reward is the scaled difference in portfolio values between consecutive timesteps. We assume that the market dynamics are not affected by our trading. There is a small $0.1\%$ transaction cost for every trade. We use real financial data scraped from Yahoo Finance \citep{yfinance} and consider the stocks in the Dow Jones index from \DTMdate{2001-01-01} to \DTMdate{2024-07-01} (see Appendix~\ref{app:stock_trading} for sample data). We split the train, validation, and test data into time periods \DTMdate{2001-01-01}-\DTMdate{2019-01-01}, \DTMdate{2019-01-01} - \DTMdate{2021-01-01}, and \DTMdate{2021-01-01}-\DTMdate{2024-07-01}, respectively. Unlike \cite{liu2018practical}, who used only the current timestep, we use a sliding window approach and use the previous $9$ timesteps for the state. See Appendix~\ref{app:stock_trading} for a more detailed description.

\paragraph{Models} For this domain, we use SAC \citep{haarnoja2018soft} as our RL algorithm and consider equivariance to both the translation group and scale-translation group across the time dimension. \edit{Temporal translations can be useful as the most recent history of stock prices inform your actions, and this information may be approximately preserved across time. Temporal scaling could also be beneficial as there could be market seasonality, which is only approximately shared across different time scales.} As our actions do not affect stock prices, which in turn is directly correlated with the reward, we learn an approximately invariant policy and invariant critic for both symmetry groups. As before we compare approximately equivariant, strictly equivariant, and unconstrained models. We evaluate each method on the final portfolio value (equivalent to the total episode reward), annualized return, and the Sharpe ratio \citep{sharpe1994sharpe}, which is a standard financial metric that measures an asset's risk-adjusted performance. We also include as baselines a uniform holding strategy \texttt{Uniform}, where we initially buy equal values of each stock and hold, and the Dow Jones index \texttt{\textasciicircum DJI}.

\paragraph{Results}
\begin{table}[t]
\centering
\caption{Test results on the stock trading dataset. Gray values indicate $95\%$ CI over $5$ runs. The approximately equivariant agents for both scale-translation (ST) and translation (T) outperform the exactly equivariant and non equivariant methods.}
\resizebox{\columnwidth}{!}{%
\begin{tabular}{@{}llrrr@{}}
\toprule
 &  & \multicolumn{1}{c}{\begin{tabular}[c]{@{}c@{}}Final Portfolio \\ Value (\$mm)\end{tabular}} & \multicolumn{1}{c}{\begin{tabular}[c]{@{}c@{}}Annualized \\ Return (\%)\end{tabular}} & \multicolumn{1}{c}{Sharpe Ratio} \\ \midrule
\multirow{2}{*}{ApproxEquiv} & \multicolumn{1}{l|}{ST} & $1.489 {\color{gray} \scriptstyle \pm 0.16}$ & $12.0 {\color{gray} \scriptstyle \pm 3.4}$ & $0.63 {\color{gray} \scriptstyle \pm 0.1}$ \\
 & \multicolumn{1}{l|}{T} & $1.428 {\color{gray} \scriptstyle \pm 0.04}$ & $10.6 {\color{gray} \scriptstyle \pm 3.8}$ & $0.60 {\color{gray} \scriptstyle \pm 0.1}$ \\
\multirow{2}{*}{ExactEquiv} & \multicolumn{1}{l|}{ST} & $1.411 {\color{gray} \scriptstyle \pm 0.15}$ & $10.3 {\color{gray} \scriptstyle \pm 3.4}$ & $0.62 {\color{gray} \scriptstyle \pm 0.2}$ \\
 & \multicolumn{1}{l|}{T} & $1.307 {\color{gray} \scriptstyle \pm 0.18}$ & $7.8 {\color{gray} \scriptstyle \pm 4.3}$ & $0.50 {\color{gray} \scriptstyle \pm 0.3}$ \\
\multicolumn{2}{l|}{NonEquiv} & $1.378 {\color{gray} \scriptstyle \pm 0.05}$ & $9.6 {\color{gray} \scriptstyle \pm 1.3}$ & $0.62 {\color{gray} \scriptstyle \pm 0.1}$ \\
\multicolumn{2}{l|}{Uniform} & $1.412$ & $10.4$ & $0.71$ \\
\multicolumn{2}{l|}{\textasciicircum DJI} & $1.293$ & $7.7$ & $0.53$ \\ \bottomrule
\end{tabular}
}
\label{tab:stock_trading}
\end{table}

\begin{figure}[t]
    \centering
    \begin{subfigure}[b]{0.49\columnwidth}
         \centering
         \includegraphics[width=\textwidth]{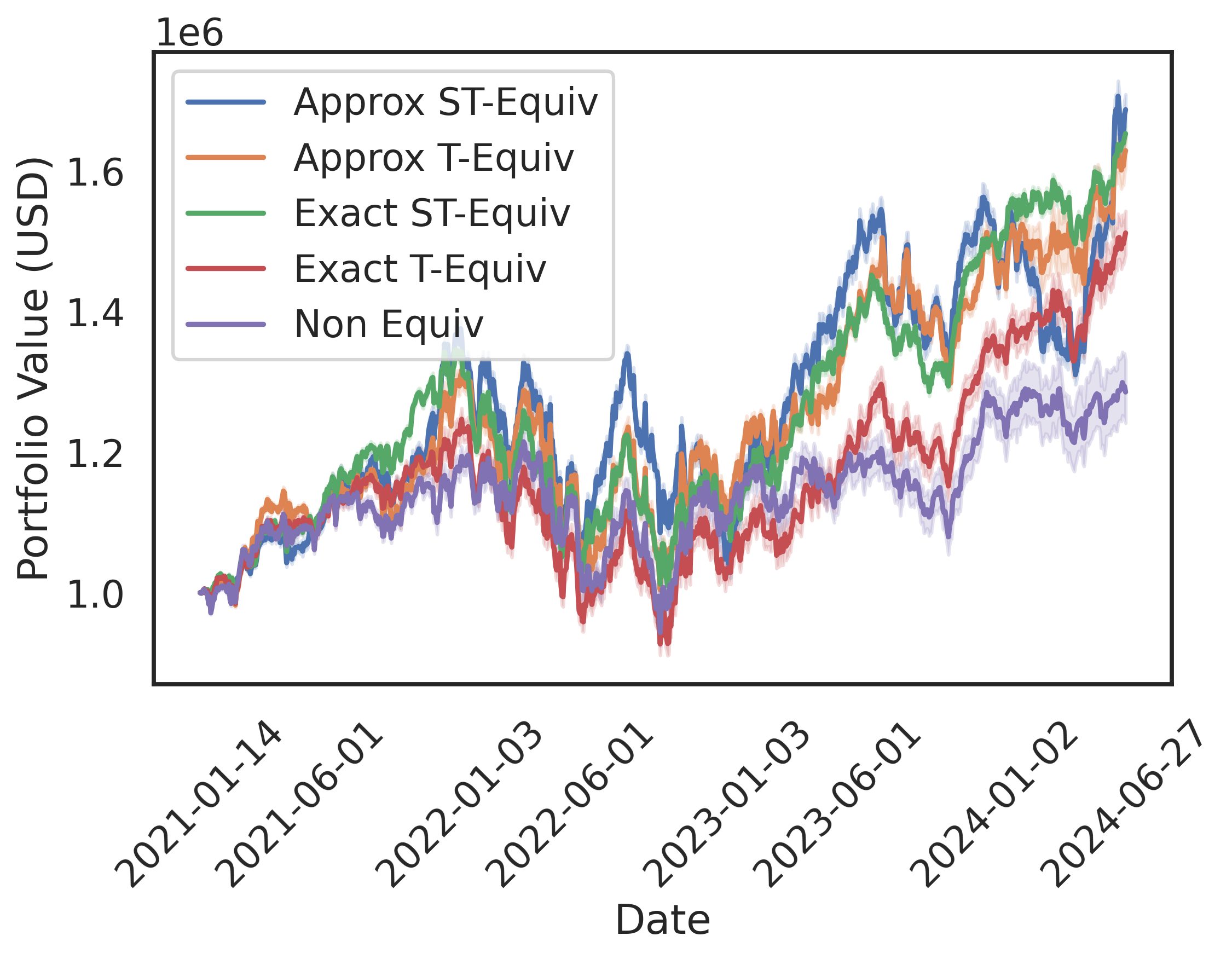}
     \end{subfigure}
    \begin{subfigure}[b]{0.49\columnwidth}
         \centering
         \includegraphics[width=\textwidth]{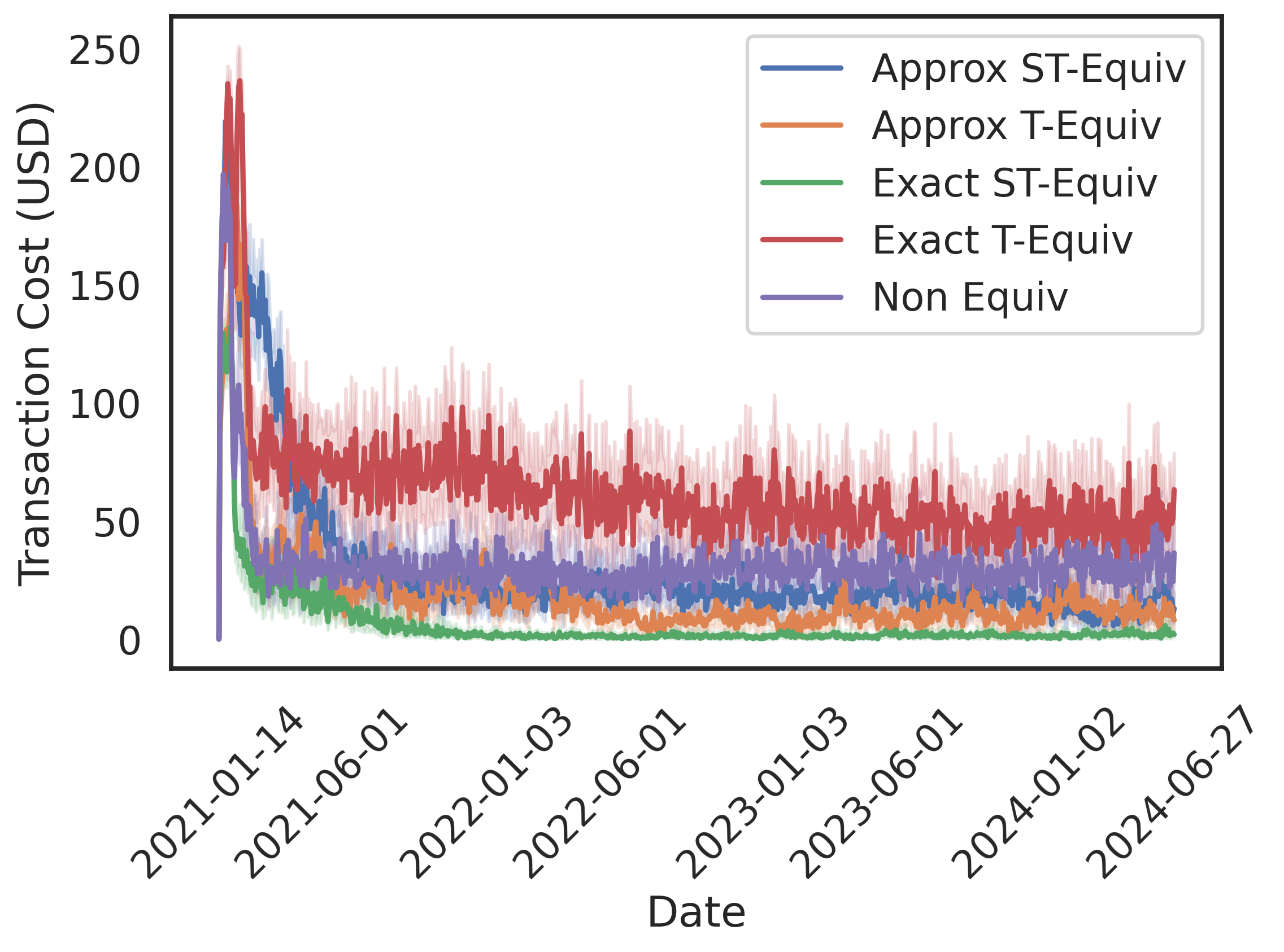}
     \end{subfigure}
    \caption{$10$ episode rollouts from the best performing policy for each method. Approx ST-Equiv often achieves the highest portfolio value for each time step and incurs minimal transaction costs.}
    \label{fig:stock_trading_rollout}
    \vspace{-0.3cm}
\end{figure}

Table~\ref{tab:stock_trading} lists the average test results of the learned policies on the stock trading domain. The \texttt{ApproxEquiv} model for both translation (T) and scale-translation (ST) outperform all baselines, with annualized returns of $10.6\%$ and $12.0\%$ respectively. The \texttt{Exact ST-Equiv} model outperforms \texttt{NonEquiv}, while the \texttt{Exact T-Equiv} model does worse. These observations suggest that temporal scale and translation symmetries can be good biases in analyzing financial data and that translation symmetry may be more approximate than scale. We also visualize $10$ episode rollouts of the best-performing policies in Figure~\ref{fig:stock_trading_rollout}, with the portfolio values on the left and transaction costs on the right. The Approx ST-Equiv method achieves the highest portfolio value for most timesteps and incurs lower transaction costs than the exactly equivariant policies. We note that overall the annualized returns are fairly low, as the test dataset from \DTMdate{2021-01-01} to \DTMdate{2024-07-01} includes both the COVID-19 pandemic and 2022 stock market decline. 

\begin{figure}[t]
    \centering
    \begin{subfigure}[b]{0.49\columnwidth}
         \centering
         \includegraphics[width=\textwidth]{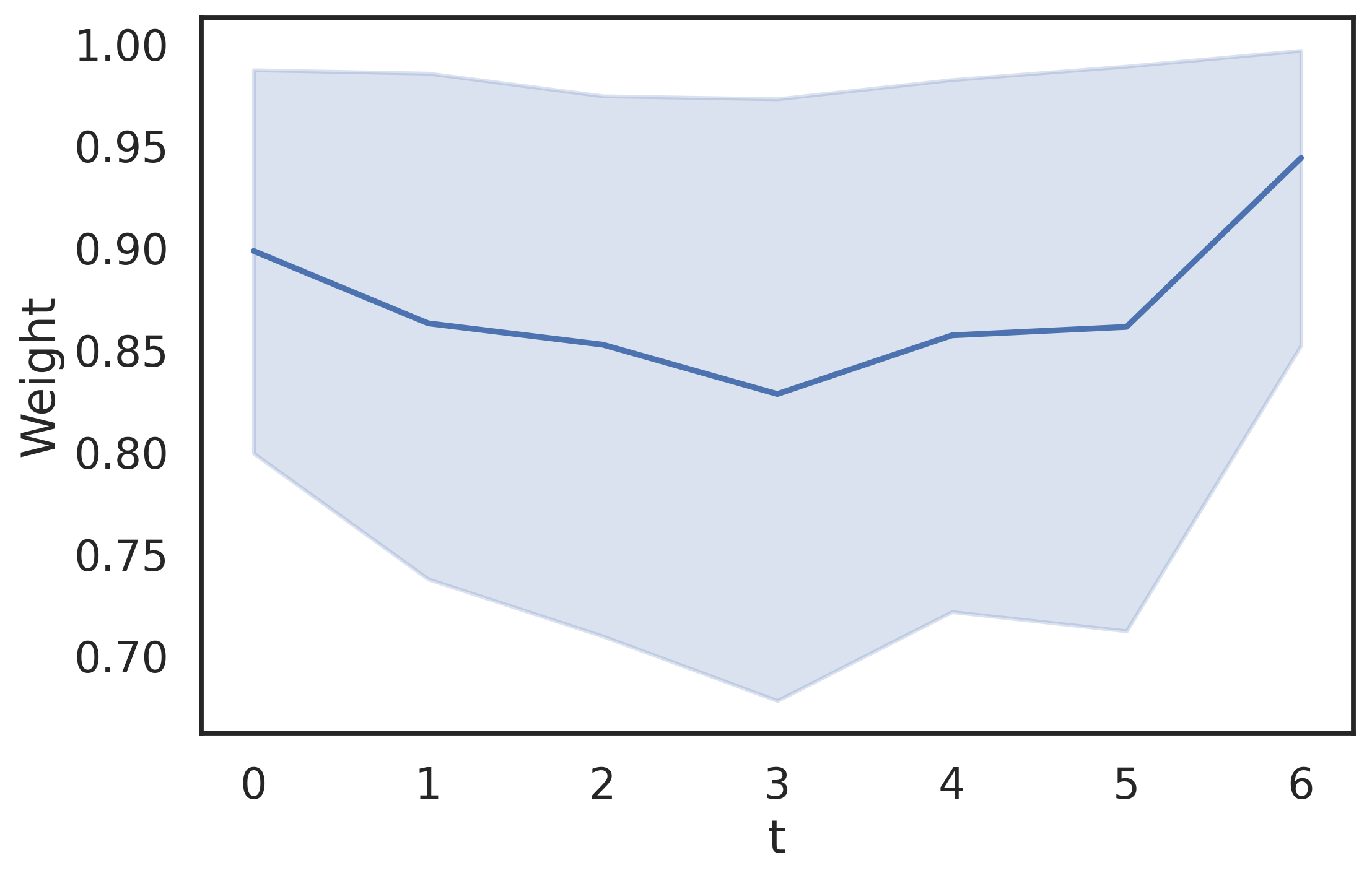}
     \end{subfigure}
    \begin{subfigure}[b]{0.49\columnwidth}
         \centering
         \includegraphics[width=\textwidth]{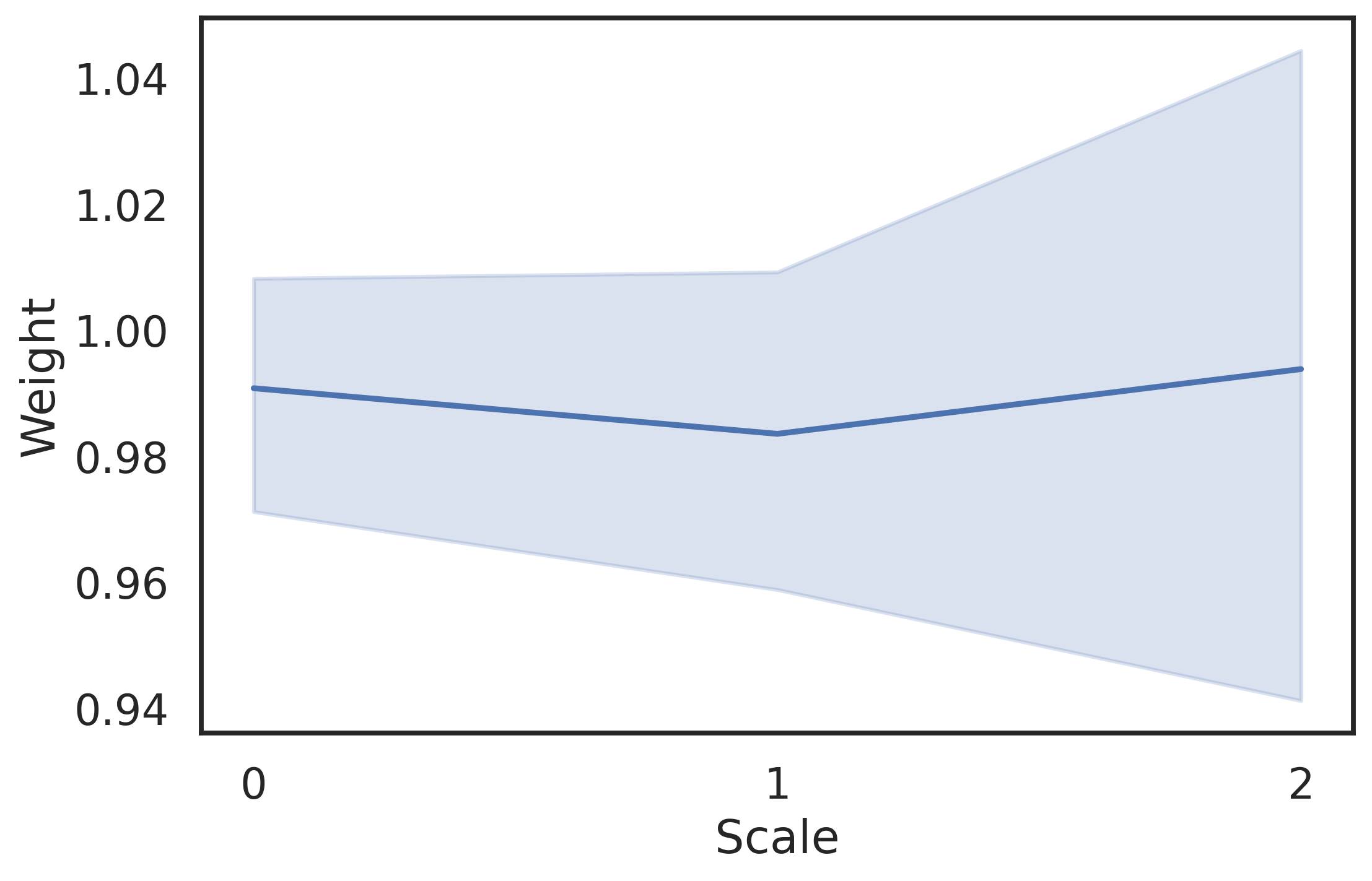}
     \end{subfigure}
    \caption{Visualization of relaxed weights for translation (left) and scale (right) over all runs. The relaxed weights for translation differ for each timestep and are similar for scale.}
    \label{fig:stock_trading_weights}
    \vspace{-0.3cm}
\end{figure}
We visualize the relaxed weights of the first layer of the encoder across translation (left) and scale (right) in Figure~\ref{fig:stock_trading_weights}. For translation, our model places higher weights on the very last timestep. This matches our intuition as the most recent stock prices and portfolio holdings would be most informative in determining the optimal action. For scale, we find that the relaxed weights do not differ greatly, but there is increased variance with increasing scale.

\section{DISCUSSION}

We proposed a novel approximately equivariant architecture using relaxed group convolutions for model-free reinforcement learning. 
%
Our experimental results on continuous control domains and a stock trading problem with real-world data demonstrate that the approximately equivariant model performs similarly to an exactly equivariant model in domains with perfect symmetry but outperforms it in most domains with symmetry breaking factors. This suggests that our method can act as a much more flexible alternative to exactly equivariant agents that can boost sample efficiency in a wider variety of settings and is also more robust to perturbations.

\paragraph{Limitations and Future Work} While we did consider real-world data in the stock trading domain, our continuous control domains used simplified observations and synthetic symmetry breaking.    
Furthermore, exactly equivariant networks perform better in some modified domains than others (\texttt{Reacher} vs. \texttt{Acrobot}). Another limitation is that, as with all equivariant networks, the symmetry group and how it acts on the state and action spaces need to be known in advance.
An interesting future direction could be to quantify exactly what types of symmetry breaking factors could lead to higher performance for approximately equivariant RL, possibly by measuring equivariance error. 
Other future work includes proving bounds on the optimal policy $\pi(s)$ and $\pi(gs)$ or applying approximately equivariant RL in robotic manipulation, where kinematic constraints or obstacles can break symmetry.

{\sujay{Future Work: Need to know the symmetry groups.. how to deal with this.. generalization..evaluation metrics }}

\edit{
\subsection*{Acknowledgments}
This project was supported in part by NSF grants \#2134178, 2107256, 2314182. The authors thank Hyunwoo Ryu for helpful discussions and pointing us to the Residual Pathways Prior baseline.
}

\subsubsection*{Disclaimer}
This paper was prepared for informational purposes [“in part” if the work is collaborative with
external partners] by the Artificial Intelligence Research group of JPMorgan Chase \& Co. and its
affiliates (``JP Morgan'') and is not a product of the Research Department of JP Morgan. JP Morgan
makes no representation and warranty whatsoever and disclaims all liability, for the completeness,
accuracy or reliability of the information contained herein. This document is not intended as
investment research or investment advice, or a recommendation, offer or solicitation for the purchase
or sale of any security, financial instrument, financial product or service, or to be used in any way
for evaluating the merits of participating in any transaction, and shall not constitute a solicitation
under any jurisdiction or to any person, if such solicitation under such jurisdiction or to such person
would be unlawful.

\bibliography{references}

\newpage
\section*{Checklist}

 \begin{enumerate}

 \item For all models and algorithms presented, check if you include:
 \begin{enumerate}
   \item A clear description of the mathematical setting, assumptions, algorithm, and/or model. [Yes]
   \item An analysis of the properties and complexity (time, space, sample size) of any algorithm. [No]
   \item (Optional) Anonymized source code, with specification of all dependencies, including external libraries. [Yes]
 \end{enumerate}

 \item For any theoretical claim, check if you include:
 \begin{enumerate}
   \item Statements of the full set of assumptions of all theoretical results. [Yes]
   \item Complete proofs of all theoretical results. [Yes]
   \item Clear explanations of any assumptions. [Yes]
 \end{enumerate}

 \item For all figures and tables that present empirical results, check if you include:
 \begin{enumerate}
   \item The code, data, and instructions needed to reproduce the main experimental results (either in the supplemental material or as a URL). [Yes]
   \item All the training details (e.g., data splits, hyperparameters, how they were chosen). [Yes]
         \item A clear definition of the specific measure or statistics and error bars (e.g., with respect to the random seed after running experiments multiple times). [Yes]
         \item A description of the computing infrastructure used. (e.g., type of GPUs, internal cluster, or cloud provider). [Yes]
 \end{enumerate}

 \item If you are using existing assets (e.g., code, data, models) or curating/releasing new assets, check if you include:
 \begin{enumerate}
   \item Citations of the creator If your work uses existing assets. [Yes]
   \item The license information of the assets, if applicable. [Yes]
   \item New assets either in the supplemental material or as a URL, if applicable. [Yes]
   \item Information about consent from data providers/curators. [No]
   \item Discussion of sensible content if applicable, e.g., personally identifiable information or offensive content. [Not Applicable]
 \end{enumerate}

 \item If you used crowdsourcing or conducted research with human subjects, check if you include:
 \begin{enumerate}
   \item The full text of instructions given to participants and screenshots. [Not Applicable]
   \item Descriptions of potential participant risks, with links to Institutional Review Board (IRB) approvals if applicable. [Not Applicable]
   \item The estimated hourly wage paid to participants and the total amount spent on participant compensation. [Not Applicable]
 \end{enumerate}

 \end{enumerate}

\appendix

\onecolumn
\aistatstitle{Approximate Equivariance in Reinforcement Learning: \\ Supplementary Materials}

\setcounter{section}{0}
\setcounter{theorem}{0}
\allowdisplaybreaks

\section{PROOF OF THEOREM~\ref{thm:Gap_Q}}
\label{app:proofs}
The proof is established by first deriving the deviation for an (arbitrary) finite-horizon discounted problem and then using this to derive the bounds for the infinite horizon case. All intermediate results are collected as propositions.

Consider an discounted-reward finite-horizon (horizon length is $T$ rather than infinity) MDP with the same reward and transition kernel as the original MDP (independent of time).
For a given stochastic policy~$\pi = (\pi_1,\pi_2\cdots,\pi_{T-1})$, let
\begin{align*}
    \mathcal{V}_t^{\pi}(s) &= \mathbb{E}^{\pi} \Big[ \sum_{k=t}^{T-1} \gamma^{k-t} R(s_k,a_k) \Big| s_t = s \Big], \\
    \mathcal{Q}^{\pi}_t(s,a) &= \mathbb{E}^{s_{t+1}} \Big[R(s,a) + \gamma \mathcal{V}^{\pi}_{t+1}(s_{t+1}) \Big| s_{t} = a, a_t = a \Big],
\end{align*}
be the finite-horizon counterparts of the expected return and action-value.
Recursively define the policy independent counterparts as follows: 
\begin{align*}
  \mathcal{V}_{T}(s_{T}) &=0,~~  \mathcal{V}_{T}(gs_{T}) = 0,\\
    \mathcal{Q}_t(s_t,a_t) &= R(s_t,a_t) +  \gamma \int_{\mathcal{S}} \mathcal{V}_{t+1}(s_{t+1})P(s_{t+1}|s_t,g_t), \\
    \mathcal{Q}_t(gs_t,ga_t) & = R(gs_t,ga_t) + \gamma \int_{\mathcal{S}}\mathcal{V}_{t+1}(gs_{t+1}) P(gs_{t+1}|gs_t,ga_t), \\
    \mathcal{V}_t(s_t) &=   \sup_{a_t \in \mathcal{A}} \mathcal{Q}_t(s_t,a_t),~~
    \mathcal{V}_t(gs_t) = \sup_{a_t \in \mathcal{A}} \mathcal{Q}_t(gs_t,ga_t).
\end{align*}

{
We also define
\begin{align*}
V^{\pi}_{t}(s) := \mathbb{E}^{\pi}\Big[ \sum_{k=t}^{\infty} \gamma^{k-t} R(s_k,a_k) \Big| s_t = s \Big],\quad
Q_t^{\pi}(s,a) := \mathbb{E}^{\pi}\Big[ \sum_{k=t}^{\infty} \gamma^{k-t} R(s_k,a_k) \Big| s_t = s, a_t = a \Big].
\end{align*}
Let~$V_t(s_t):= \sup_{\pi}V^{\pi}_t(s_t)$ and 
\[
Q_t(s_t,a_t) = \mathbb{E}\Big[R(s_a,a_t) + \gamma V_{t+1}(s_{t+1}) \Big| s_t,a_t \Big].
\]
Note that $V_t^{\pi}$ and $Q_t^{\pi}$ equal $V^{\pi}$ and $Q^{\pi}$ defined in \eqref{eq:def_V_Q} for any $t$, and $V_t$ and $Q_t$ equal $V^{*}$ and $Q^{*}$. We introduce the notations only to make the connection between the finite-horizon and infinite-horizon MDPs clearer.
}


\begin{proposition} \label{prop:q_bound}
For a $(G, \epsilon_R, \epsilon_P)$-invariant MDP, the following holds at any~$t$,
\[
|\mathcal{Q}_t(s_t,a_t) - \mathcal{Q}_t(gs_t,ga_t)| \leq \alpha_t,~~\text{and}~~ |\mathcal{V}_t(s_t) - \mathcal{V}_t(gs_t)| \leq \alpha_t,
\]
where~$\alpha_t$ is given by the following recursion: $\alpha_{T+1}=0$ and 
\[
\alpha_t = \epsilon_R + \gamma \Big\{ \rho_{\mathscr{F}}(\mathcal{V}_{t+1}) \epsilon_{P} + \alpha_{t+1} \Big\}.
\]
\end{proposition}
\vspace{\fill}

\begin{proof}
We will prove the results using induction. First, note that the result is true for~$T$ by definition. Suppose the result is true for~$t+1$, and consider the differential at time~$t$,
\begin{align*}
  |\mathcal{Q}_t(s_t,a_t) - \mathcal{Q}_t(gs_t,ga_t)| &\leq |R(s_t,a_t)-R(gs_t,ga_t)| \\ 
  &+ \gamma \Big|\int_{\mathcal{S}} {\mathcal{V}}_{t+1}(s_{t+1}) P(s_{t+1}|s_t,a_t) - \int_{\mathcal{S}} \mathcal{V}_{t+1}(gs_{t+1}) P(gs_{t+1}|gs_t,ga_t) \Big| \\
  &\leq \epsilon_R + \gamma \Big|\int_{\mathcal{S}} {\mathcal{V}}_{t+1}(s_{t+1}) P(s_{t+1}|s_t,a_t) - \int_{\mathcal{S}} \mathcal{V}_{t+1}(gs_{t+1}) P(s_{t+1}|s_t,a_t) \Big| \\
  &+ \gamma \Big|\int_{\mathcal{S}} \mathcal{V}_{t+1}(gs_{t+1}) P(s_{t+1}|s_t,a_t) - \int_{\mathcal{S}} \mathcal{V}_{t+1}(gs_{t+1}) P(gs_{t+1}|gs_t,ga_t)\Big| \\
  &\leq \epsilon_{R} + \gamma \rho_{\mathscr{F}}(\mathcal{V}_{t+1}) \epsilon_{P} + \gamma \int_{\mathcal{S}} \Big| \mathcal{V}_{t+1}(s_{t+1}) - \mathcal{V}_{t+1}(g s_{t+1}) \Big| P(s_{t+1}|s_t,a_t).
\end{align*}
The last inequality follows by using the decomposition using Minkowski's functional. Further, note that
\[
\Big| \mathcal{V}_{t+1}(s_{t+1}) - \mathcal{V}_{t+1}(g s_{t+1}) \Big| \leq \sup_{a_{t+1} \in \mathcal{A}} |\mathcal{Q}_{t+1}(s_{t+1},a_{t+1}) - \mathcal{Q}_{t+1}(gs_{t+1},ga_{t+1})| \leq \alpha_{t+1}, 
\]
by induction assumption, and the fact that when~$g$ is onto
\[
\sup_{a' \in g\mathcal{A}} \mathcal{Q}_{t+1}(gs_t,a') = \sup_{a \in \mathcal{A}} \mathcal{Q}_{t+1}(gs_t,ga).
\]
The result follows.
\end{proof}

\begin{proposition} \label{prop:Q_limits}
    Let the rewards $R \in [R_{\min},R_{\max}]$. For an arbitrary, but finite, horizon~$T$
\begin{align*}
    \mathcal{Q}_{t}(s_t,a_t) + \frac{\gamma^{T-t} }{1 - \gamma}R_{\min} &\leq Q_t(s_t,a_t) \leq \mathcal{Q}_{t}(s_t,a_t) + \frac{\gamma^{T-t} }{1 - \gamma}R_{\max}.
\end{align*}
\end{proposition}

\begin{proof}
We have by definition, 
    \begin{align*}
        Q_t(s_t,a_t) &= \mathbb{E} \Big[ \sum_{k=t}^{\infty} \gamma^{k-t} R(s_k,a_k) \Big|s_t = s, a_t= a \Big] \\
        &= \mathbb{E} \Bigg[ R(s_t,a_t) + \gamma \mathbb{E} \Big[ \sum_{k=t+1}^{\infty} \gamma^{k - (t+1)} R(s_k,a_k) \Big| s_{t+1} \Big] \Bigg| s_t = s, a_t= a \Bigg] \\
        &\leq \mathbb{E} \Bigg[ R(s_t,a_t) + \gamma \mathbb{E} \Big[ \mathcal{V}_{t+1}(s_{t+1}) + \frac{\gamma^{T-(t+1)} R_{\max}}{1 - \gamma} \Big]\Big| s_{t+1} \Big] \Bigg| s_t = s, a_t= a \Bigg] \\
        &= \mathcal{Q}_{t}(s_t,a_t) +  \frac{\gamma^{T-t} }{1 - \gamma}R_{\max}. 
    \end{align*}
Similarly, we have
\begin{align*}
    Q_t(s_t,a_t) &= \mathbb{E} \Big[ \sum_{k=t}^{\infty} \gamma^{k-t} R(s_k,a_k) \Big|s_t = s, a_t= a \Big] \\
    &\geq \mathbb{E} \Bigg[ \sum_{k=t}^{T-1} \gamma^{k-t} R(s_k,a_k) + \sum_{k=T}^{\infty} \gamma^{k-t} R_{\min} \Big| s_t, a_t \Bigg] \\
    &= \mathbb{E} \Bigg[ R(s_t,a_t) + \gamma \sum_{k=t+1}^{T-1} \gamma^{k-(t+1)} R(s_k,a_k) \Big| s_t, a_t \Bigg] + \frac{\gamma^{T-t} }{1 - \gamma}R_{\min} \\
    &= \mathcal{Q}_{t}(s_t,a_t) + \frac{\gamma^{T-t} }{1 - \gamma}R_{\min}. 
\end{align*}
\end{proof}

We now prove Theorem~\ref{thm:Gap_Q}. Let~$\mathscr{B}(\mathcal{S})$ denote the Banach space of bounded real-valued functions  on~$\mathcal{S}$. We define the Bellman optimality operator~$\mathcal{B}: \mathscr{B}(\mathcal{S}) \rightarrow \mathscr{B}(\mathcal{S})$ such that for any uniformly bounded function~$V \in \mathscr{B}(\mathcal{S})$,
\begin{align}
\mathcal{B}V(s)=\sup_{a \in \mathcal{A}} \Big\{R(s,a) + \gamma \int_{\mathcal{S}} V(s') P(s' | s,a) \Big\}\,\quad\forall s\in\mathcal{S}.
\end{align}
It is known that~$V^*$ is the (unique) fixed point of $\mathcal{B}$, i.e., $\mathcal{B} V^* = V^*$. We note that $V^*$ also satisfies the following equation for any $s$
\begin{equation} \label{eq:Bel}
V^*(gs) = \sup_{a \in \mathcal{A}} \Big\{R(gs,ga) + \gamma \int_{\mathcal{S}} V^*(gs') P(gs' | gs,ga) \Big\}.
\end{equation}

To see this, consider the following arguments.
\begin{align*}
    Q^*(s,a) &= R(s,a) + \gamma \sup_{a' \in \mathcal{A}} \int_{s' \in \mathcal{S}} Q^*(s,a) P(s'|s,a), \\
    Q^*(gs,ga) &= R(gs,ga) + \gamma \sup_{a' \in \mathcal{A}} \int_{s' \in \mathcal{S}} Q^*(gs,ga) P(s'|gs,ga).     
\end{align*}
Since~$g \in G$ permutes the elements of~$G$, re-indexing the integral using~$\tilde{s}' = gs'$, we have
\begin{align*}
    Q^*(gs,ga) &= R(gs,ga) + \gamma \sup_{\tilde{a} \in g\mathcal{A}} \int_{\tilde{s}' \in g\mathcal{S}} Q(\tilde{s}', \tilde{a}') P(\tilde{s}' | gs,ga). \\
    \therefore Q^*(gs,ga) &= R(gs,ga) + \gamma \sup_{a' \in \mathcal{A}} \int_{s' \in \mathcal{S}} Q(gs',ga') P(gs' | gs, ga).
\end{align*}
\subsection*{Proof of Theorem~\ref{thm:Gap_Q}:}
%
%
Consider a sequence of value functions~$\mathcal{V}^{(n)}$ on the symmetry transformed domain as follows: $\mathcal{V}^{(0)}(gs) = 0$ and $\mathcal{V}^{(n+1)} = \mathcal{B} \mathcal{V}^{(n)}$. For an arbitrary~$T$, we have using Proposition~\ref{prop:q_bound} for any~$t \in \{1,\cdots,T\}$,
\[
|\mathcal{V}_t(s_t) - \mathcal{V}^{(T-t)}_t(gs_t) | \leq \alpha_t,
\]
where \[
\alpha_t = \epsilon_R + \sum_{\tau = t+1}^{T-1} \gamma^{\tau - t} [\rho_{\mathscr{F}}(\mathcal{V}^{(T-\tau)}) \epsilon_{P} + \epsilon_{R}].
\]
From Proposition~\ref{prop:Q_limits}, we have, noting that $\mathcal{V}(s) = \sup_a \mathcal{Q}(s,a)$, that
\begin{align*}
    \mathcal{V}^{(T-t)}_t(gs_t) -\alpha_t + \frac{\gamma^{T-t} }{1 - \gamma}R_{\min} &\leq V_t(s_t) \leq \mathcal{V}^{(T-t)}_t(gs_t) + \alpha_t + \frac{\gamma^{T-t} }{1 - \gamma}R_{\max}
\end{align*}
By Banach fixed point theorem, we know that~$\lim_{T \rightarrow \infty} \mathcal{V}^{(T-t)}_t = V^*$. By continuity of~$\rho_{\mathscr{F}}(\cdot)$, we have that $\lim_{T \rightarrow \infty} \rho_{\mathscr{F}}(\mathcal{V}^{(T-\tau)}) = \rho_{\mathscr{F}} (V^*)$ whence  $\lim_{T \rightarrow \infty} \alpha_t = \alpha:= \frac{\epsilon_R + \gamma \rho_{\mathscr{F}}(V^*) \epsilon_P}{1 - \gamma}$. Therefore, taking the limit, we have
\[
V^*(gs_t) - \alpha \leq V_t(s_t) \leq V^*(gs_t) + \alpha.
\]
A similar argument establishes the result for~$Q$ using the onto function~$g$. 
The claims in Theorem~\ref{thm:Gap_Q} follows by recognizing that $V_t$ and $Q_t$ exactly equal $V^{*}$ and $Q^{*}$.

\qed

\section{CASE OF FINITE HORIZON: NO DISCOUNTING} \label{sec:Ap_nodisc}
As is clear from Theorem~\ref{thm:Gap_Q}, when~$\gamma \rightarrow 1$, the bound becomes trivial and not useful. In this section we will briefly discuss the case when the discount factor~$\gamma = 1$. In this setting, we allow the transition functions to be a function of~$t$. 
\begin{proposition}
    Let~$|R(gs_t,ga_t) - R(s_t,a_t)| \leq \epsilon_R$ and $d_{\mathscr{F}} \Big( P_t(gs'_t \mid gs_t, ga_t) , P_t(s'_t \mid s_t, a_t) \Big)  \leq \epsilon_P(t)$. For a finite-horizon MDP of duration~$T$, we have 
    \[
    |\mathcal{Q}_t(s_t,a_t) - \mathcal{Q}_t(gs_t,ga_t)| \leq \alpha_t,~~ |\mathcal{V}_t(s_t) - \mathcal{V}_t(gs_t)| \leq \alpha_t
    \]
    where 
    $\alpha_{T+1}=0$ and for $t \in \{1, 2, \cdots, T \}$, 
    \[
    \alpha_t = \epsilon_R + \sum_{\tau = t+1}^T \Big[ \rho_{\mathscr{F}}(\mathcal{V}_{\tau}) \epsilon_P(\tau-1) + \epsilon_R \Big].
    \]
\end{proposition}
\begin{proof}
    The proof proceeds as in Proposition~\ref{prop:q_bound}. We have 
    \begin{align*}
  |\mathcal{Q}_t(s_t,a_t) - \mathcal{Q}_t(gs_t,ga_t)| &\leq |R(s_t,a_t)-R(gs_t,ga_t)| \\ 
  &+  \Big|\int_{\mathcal{S}} {\mathcal{V}}_{t+1}(s_{t+1}) P_{t}(s_{t+1}|s_t,a_t) - \int_{\mathcal{S}} \mathcal{V}_{t+1}(gs_{t+1}) P_{t}(gs_{t+1}|gs_t,ga_t) \Big| \\
  &\leq \epsilon_R +  \Big|\int_{\mathcal{S}} {\mathcal{V}}_{t+1}(s_{t+1}) P_{t}(s_{t+1}|s_t,a_t) - \int_{\mathcal{S}} \mathcal{V}_{t+1}(gs_{t+1}) P_{t}(s_{t+1}|s_t,a_t) \Big| \\
  &+  \Big|\int_{\mathcal{S}} \mathcal{V}_{t+1}(gs_{t+1}) P_t(s_{t+1}|s_t,a_t) - \int_{\mathcal{S}} \mathcal{V}_{t+1}(gs_{t+1}) P_t(gs_{t+1}|gs_t,ga_t)\Big|  \\
  &\leq \epsilon_R + \rho_{\mathscr{F}}(\mathcal{V}_{t+1}) \epsilon_P(t) + \alpha_{t+1} := \alpha_t.
  \end{align*} 
  The result follows by recursion. 
\end{proof}

\section{BACKGROUND AND METHOD}
\label{app:method}

\subsection{Equivariance with Group Convolutions}
\label{app:group_conv}
Group convolutions \citep{cohen2016group} generalize standard convolutions, which are translation-equivariant, to be equivariant to a group $G$. Group convolutions act on signals over the group $f: G \rightarrow \mathbb{R}$. As many data samples are not natively of this form (e.g. an image), the input must first be lifted onto a function in $G$. For example, let $f_0: \mathbb{Z}^2 \rightarrow \mathbb{R}$ be the input signal, a grayscale image, and $H=D_2$ be the group. The lifting convolution lifts $f_0$ from $\mathbb{Z}^2$ to $G=D_2 \ltimes \mathbb{Z}^2$ by
\begin{align}
    (f_0 \star \psi)(x, h) = \sum_{y \in \mathbb{Z}^2} f_0(y) \psi(h^{-1}(y - x)),
\end{align}
where $h \in H$. Practically, the lift operation creates $|H|$, the order of group $H$, images by acting on $x$ by $h^{-1}$. Typically the lift operation is the first layer of the network, followed by subsequent group convolutions, nonlinearities, or other equivariant layers. We use relaxed versions of the lift and group convolutions as described in \cite{wang2022approximately} and the main paper.

\subsection{Steerable Convolutions}
\label{app:steerable_conv}
As an alternative to group convolutions, one can use steerable convolutions \citep{weiler20183d} that use weight tying to generalize to continuous groups and are more parameter-efficient. Let $H < O(2)$ be the subgroup which acts on $\mathbb{R}^2$ by matrix multiplication on the input and output channel spaces $\mathbb{R}^c$ and $\mathbb{R}^d$ by $\rho_{\textrm{in}}$ and $\rho_{\textrm{out}}$, respectively. Then $G = H \ltimes \mathbb{R}^2$. Given input signal $f:\mathbb{R}^2 \rightarrow \mathbb{R}^{c}$, then standard convolution over $\mathbb{R}^2$ with kernel $\psi:\mathbb{R}^2 \rightarrow \mathbb{R}^{d \times c}$ is $G$-equivariant if $\psi$ satisfies
\begin{align}
\label{eq:equiv_constraint}
    \psi(hx) = \rho_{\textrm{out}}(g)\psi(x)\rho_{\textrm{in}}(h^{-1}), 
\end{align}
for all $h \in H$. Intuitively, this kernel constraint ensures that the output features transform by $\rho_{\textrm{out}}$ when the input features are transformed by $\rho_{\textrm{in}}$. Kernels that satisfy this constraint have been solved for many common subgroups of $E(2)$, see \cite{weiler2019general} for more details.

Using the example of grayscale images as in Section~\ref{app:group_conv}, let the input feature be $f: \mathbb{Z}^2 \rightarrow \mathbb{R}$ and $\{\psi_k\}_{k=1}^{K}$ be a precomputed, nontrainable equivariant kernel basis of $K$ kernels that satisfy Eq.~\eqref{eq:equiv_constraint}. Assume that both the number of input and output channels is $1$ and let $w \in \mathbb{R}^{K}$ be the trainable coefficients of the kernels. Then a $G$-steerable convolution is defined as
\begin{align}
    (f \star \psi)(x) = \sum_{y \in \mathbb{Z}^2} \sum_{k=1}^{K} (w_k \psi_k(y))f(x + y),
\end{align}
where $x \in \mathbb{Z}^2$ is the spatial position and $w_k$ is the weight associated with kernel $\psi_k$. 

\paragraph{Relaxed Steerable Convolution}
As described in \cite{wang2022approximately}, one can also use relaxed versions of steerable convolutions by letting the trainable weights $w$ also depend on $y$. A relaxed $G$-steerable convolution is defined as
\begin{align}
    (f \tilde{\star} \psi)(x) = \sum_{y \in \mathbb{Z}^2} \sum_{k=1}^{K} (w_k(y) \psi_k(y)) f(x+y).
\end{align}
Allowing the trainable weights $w_k$ to also depend on the absolute spatial position $y$ breaks the equivariance constraint in Eq.~\eqref{eq:equiv_constraint}.

\begin{figure*}[t]
    \centering
    \includegraphics[trim={0 100 0 0},clip, width=0.9\textwidth]{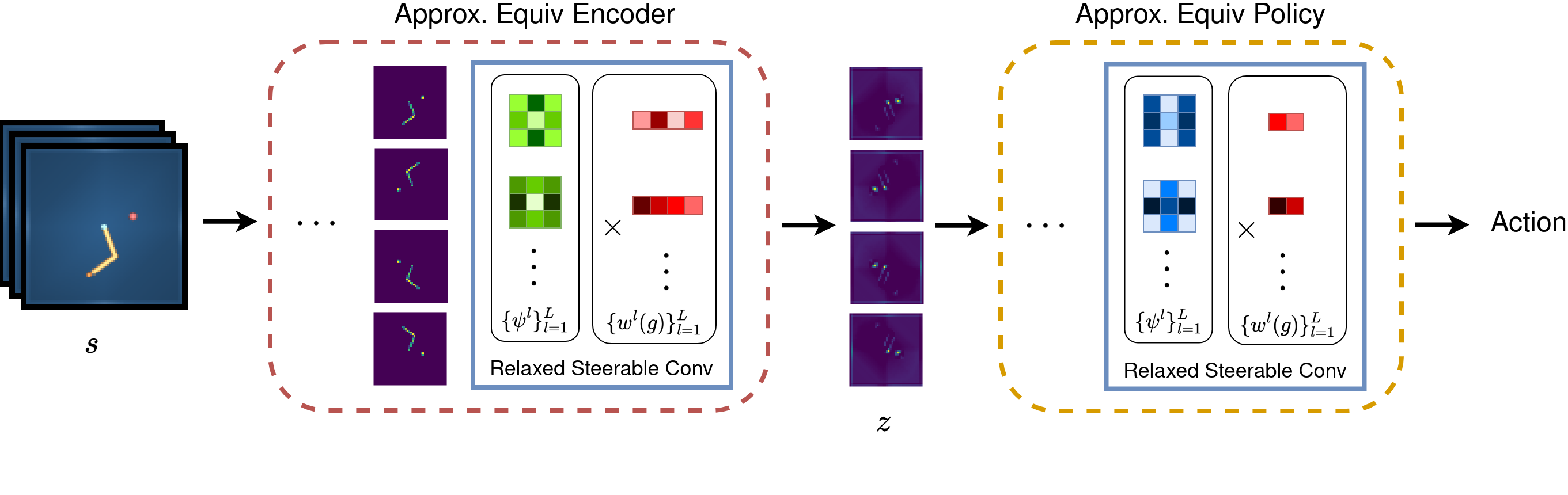}
    \caption{Illustration of an approximately $D_2$-equivariant encoder and policy using relaxed steerable convolution layers. The critic is not shown and is approximately invariant.}
    \label{fig:steerable_architecture}
\end{figure*}

By replacing relaxed group convolutions with relaxed steerable convolutions, we can also design a variant of our proposed approximately equivariant RL architecture (Figure~\ref{fig:steerable_architecture}). 

\section{EXPERIMENT DETAILS}
\label{app:exp_details}

\subsection{Continuous Control}
\label{app:dmc}

\paragraph{Acrobot}
We use the \texttt{swingup} task. The domain consists of two joints where the goal is to apply torque to the inner joint so that both joints are near vertical. We use $D_1$ as the symmetry group, i.e. vertical reflection, and the action $a\in \mathbb{R}$ transforms via the sign representation $\rho_{\textrm{sign}}$, where $\rho_{\textrm{sign}}(\textrm{flip})(a) = -a$. For variants, we consider 1) \texttt{repeat\_action} - the action is repeated when the inner joint is in the fourth quadrant and 2) \texttt{gravity} - gravity $\Vec{g}=[0,0,-9.81]$ is modified to $[-2, 2, -9.81]$.

\paragraph{CartPole}
We consider the \texttt{swingup} task. The domain consists of a pole swinging on a cart and the goal is to move the cart left or right ($a \in \mathbb{R}$) to make the pole upright. The symmetry group and action representation are the same as in \texttt{Acrobot}, $D_1$ and $\rho_{\textrm{sign}}$. For variants, we consider 1) \texttt{repeat\_action} - the action is repeated when the pole is in the first quadrant, 2) \texttt{gravity} - gravity is modified to $[0.2, -0.2, -9.81]$, and 3) \texttt{reflect\_action} - the pole angle is in $[0, \frac{\pi}{4}]$. Gravity is modified less than in \texttt{Acrobot} as too high values forced the cart out of frame.

\paragraph{Cup Catch}
The domain consists of a ball attached to the bottom of the cup and the goal is to move the cup to catch the ball inside the cup. The action $(x,z) \in \mathbb{R}^2$ is the cup's spatial position. The symmetry group is $D_1$ and the action representation is $\rho_{\textrm{sign}} \oplus \rho_0$, where the $x$ position transforms via the sign representation and the $z$ transforms via the trivial representation $\rho_0$. For variants, we consider 1) \texttt{repeat\_action} - the ball $x$ position greater than $0.0$ and $z$ position is greater than $0.3$, 2) \texttt{gravity} - gravity is modified to $[-2, 2, -9.81]$, and 3) \texttt{reflect\_action} - same as \texttt{repeat\_action}.

\paragraph{Reacher}
We consider the \texttt{hard} task. The domain consists of two joints and the goal is to apply torques to make the end effector reach the target. The action $a \in \mathbb{R}^2$. The symmetry group is $D_2$, i.e. vertical reflections and $\pi$ rotations, and the action transforms via the quotient representation $2 \rho_{\textrm{quot}}$, where the torques for both joints are invariant to rotations and flip signs for vertical reflections. For variants, we consider 1) \texttt{repeat\_action} - the inner joint angle is in $[0, \frac{\pi}{2}]$ and 2) \texttt{reflect\_action} - the inner joint angle is in $[\frac{\pi}{2}, \pi]$.

\subsubsection{Training Details}
For all DeepMind Control Suite (DMC) domains, we fix the episode length to $1000$ and use RGB image of size $85 \times 85$. We considered four domains of varying difficulty, of which \texttt{Acrobot} is the hardest. In the original DrQv2 implementation \citep{yarats2021mastering}, the encoder reduces the spatial dimensions to $35 \times 35$, which is then flattened to be input to the policy and critic. We follow \cite{wang2022surprising} and further reduce the spatial dimensions to $7 \times 7$ for faster training for all models. We reduce the replay buffer size from $1{,}000{,}000$ to $500{,}000$ to slightly reduce the memory footprint. All other hyperparameters are kept the same as in \cite{yarats2021mastering}.

For the exactly equivariant and approximately equivariant models, we reduce the number of channels by $\sqrt{|G|}$ where $|G|$ is the order of the group to preserve roughly the same number of parameters as the non-equivariant model. We use $L=1$ filters for the approximately equivariant model in all experiments.

RPP contains both the non-equivariant layers and exactly equivariant layers and thus has roughly twice as many parameters as \texttt{ExactEquiv}. For the critic moving average speed $\tau$, we use the default $\tau=0.01$ for \texttt{CartPole} and \texttt{Reacher} and $\tau=0.009$ for \texttt{Acrobot} and \texttt{Ball in Cup}.

The plots in Figure~\ref{fig:dmc_curves} show the mean reward of $10$ episodes, evaluated every $20{,}000$ environment steps. For the results in Table~\ref{tab:dmc_test}, we use $\sigma=0.02$ for \texttt{Acrobot} and \texttt{Reacher} and $\sigma=0.06$ for \texttt{CartPole} and \texttt{Ball in Cup}.

The continuous control experiments were run on single GPUs of different types. \texttt{Acrobot} was run on an Nvidia RTX 4090 and all other experiments were run on an Nvidia RTX 2080 Ti. We note that the wall clock time for training both exactly and approximately equivariant models is longer than that for a non equivariant model, even though they are generally more sample efficient. This is because equivariant neural networks often incur more overhead in implementation - for group convolutions, the kernel must be transformed and the outputs must be stacked and for steerable convolutions, the basis must be projected onto matrices at every forward pass.

\subsection{Stock Trading}
\label{app:stock_trading}
We formulate the stock trading problem as an MDP as described in \cite{liu2018practical}. The state consists of the cash balance $c_t$, the stock prices $p^n_t$, the number of shares in the current portfolio $h^n_t$, and other technical indicators $i^n_t$ for time $t$ stock $n \in \{1,\dots,N\}$. The actions $x^n_t$ are the number of stocks to buy and sell for each stock $n$ and are bounded to $[-M, M]$ where $M$ was set to $100$. The reward $r_t$ is the scaled difference in portfolio values between consecutive timesteps and we assume that the market dynamics are not affected by our trading. There is a small transaction cost $\epsilon^n=0.001$ for every trade. Initially, the portfolio contains $0$ shares and the cash balance is $1{,}000{,}000$. This can be formulated as a constrained program as follows
\begin{align*}
    \text{max} & \quad \sum_{t} r_t \\
    \text{s.t.} & \quad -M \leq a_t^n \leq M, & \forall n, t \\
    & \quad a_t^n \geq -h_t^n, & \forall n, t \\
    & \quad a_t^n \leq \left\lfloor{c_t / (p_t^n (2 + \epsilon^n))}\right \rfloor & \forall n, t \\
    & \quad c_t \geq 0 & \forall t \\
    & \quad c_{t+1} = c_t - \sum_{n} a_t^n p_t^n (1+\epsilon^n) & \forall t\\
    & \quad h_{t+1}^n = h_t^n + a_t^n & \forall n, t \\
    & \quad r_{t+1} = (c_{t+1} - c_t) + \sum_{n} (p_{t+1}^n h_{t+1}^n - p_t^n h_t^n) & \forall t \\
    & \quad c_0 = 1{,}000{,}000 & \\
    & \quad h_0^n = 0 & \forall n \\
    & \quad h_t^n \in \mathbb{Z}^+, a_t^n \in \mathbb{Z}, c_t \in \mathbb{R}^+.
\end{align*}

\begin{figure}[htpb]
    \centering
        \includegraphics[trim={0 0 0 0}, clip, width=0.6\columnwidth]{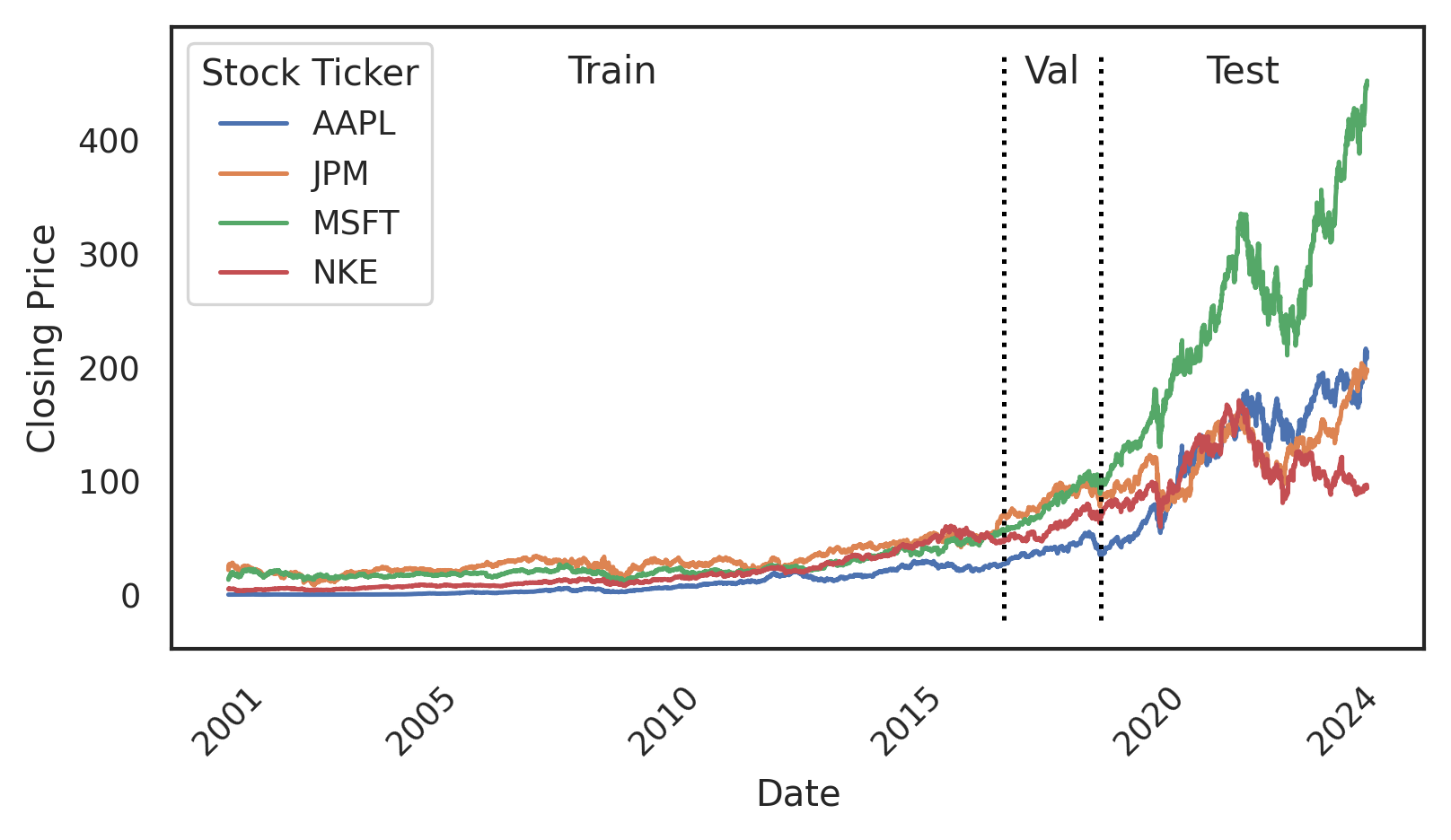}
    \caption{Sample stock trading data. We use a sliding window of the stock prices, current portfolio, cash balance, and other indicators as the state. The dataset is split into train/val/test as shown.}
    \label{fig:stock_trading_domain}
\end{figure}

The financial data was pulled from Yahoo Finance \citep{yfinance} for the time period  \DTMdate{2001-01-01} to \DTMdate{2024-07-01} (see Figure~\ref{fig:stock_trading_domain} for sample stock prices). As historical stock prices and portfolio can be important for determining the action, we use the previous $H=9$ timesteps for the state, unlike \cite{liu2018practical}.

\subsubsection{Training Details}
or all models, we use $4$ layers for the shared encoder, $1$ layer for the actor, and $2$ layers for the critic. The non equivariant model uses linear layers after flattening the input, while the exactly equivariant and approximately equivariant models use group convolutions and relaxed group convolutions with a kernel size of $5$, respectively. We consider both temporal translations and temporal scale-translations. For scale-translation, we use separable group convolutions \citep{knigge2022exploiting} and use 3 scale factors $0.8, 0.98, 1.2$. We control the number of channels so that the total number of parameters is roughly equal to the non equivariant model. We use $L=1$ filters for the approximately equivariant model in all experiments.

The stock trading experiments were run on a single Nvidia RTX 2080 Ti. All other hyperparameters are given in Table~\ref{tab:stock_params}.

\begin{table}[htpb]
\centering
\caption{Hyperparameters used for stock trading experiments}
\begin{tabular}{l|cll}
\toprule
\multicolumn{1}{c|}{Hyperparameter} & \multicolumn{1}{l}{ApproxEquiv} & ExactEquiv & NonEquiv \\ \midrule
Batch size & \multicolumn{3}{c}{64} \\
Learning rate & \multicolumn{3}{c}{1e-4} \\
$\alpha$ & \multicolumn{3}{c}{0.05} \\
$\tau$ & \multicolumn{3}{c}{0.005} \\
Discount factor & \multicolumn{3}{c}{0.99} \\
Hidden dim/channels & 64 & \multicolumn{1}{c}{64} & \multicolumn{1}{c}{128} \\
Encoder output dim/channels & \multicolumn{3}{c}{256} \\ \bottomrule
\end{tabular}
\label{tab:stock_params}
\end{table}

\edit{
\section{\edit{ROBUSTNESS WITH NOISE AUGMENTATION}}
\label{app:experiments}
Table~\ref{tab:noisy_training} shows the results from training policies with noisy inputs and evaluating their robustness to noise at test time. This experiment tests whether our approximately equivariant method is truly more robust to noise than other baselines trained with noise augmentation. We find that our approximately equivariant method is more robust than baselines for the modified domain, even when trained with noise augmentation.

\begin{table}[t]
\centering
\caption{Total episode reward on $50$ rollouts for the best policy when trained on noisy inputs and tested in the noisy domain. Gray values indicate $95\%$ CI. \texttt{ApproxEquiv} learns a more robust policy than baselines on the modified \texttt{BallInCup} domain.}
\begin{tabular}{@{}cl|rrr@{}}
 &  & \multicolumn{1}{c}{ApproxEquiv} & \multicolumn{1}{c}{ExactEquiv} & \multicolumn{1}{c}{NonEquiv} \\ \midrule
\multirow{2}{*}{\begin{tabular}[c]{@{}c@{}}\sc{Ball in Cup} \\ (Noisy)\end{tabular}} & Original & $971 {\color{gray}\scriptstyle \pm 1.7}$ & $\textbf{977} {\color{gray}\scriptstyle \pm 3.8}$ & $914 {\color{gray}\scriptstyle \pm 7.9}$ \\
 & Gravity & $\textbf{973} {\color{gray}\scriptstyle \pm 2.7}$ & $952 {\color{gray}\scriptstyle \pm 5.7}$ & $942 {\color{gray}\scriptstyle \pm 10.}$ \\ \bottomrule
\end{tabular}
\label{tab:noisy_training}
\end{table}
}

\end{document}